\pdfoutput=1

\documentclass[11pt]{article}

\usepackage{ACL2023}

\usepackage{times}
\usepackage{latexsym}

\usepackage[T1]{fontenc}

\usepackage[utf8]{inputenc}

\usepackage{microtype}

\usepackage{inconsolata}

\usepackage{amsmath}
\usepackage{amssymb}
\usepackage{mathrsfs}
\usepackage{xcolor}
\usepackage{amsthm}
\usepackage{amsfonts}
\usepackage{graphicx}
\usepackage{stmaryrd}
\usepackage{algorithm}
\usepackage{xr}
\usepackage{wrapfig}

\usepackage{latexsym}
\usepackage{url}
\usepackage{array}
\usepackage{booktabs}
\usepackage{balance}
\usepackage{colortbl}
\usepackage{makecell}
\usepackage{soul}
\usepackage{blindtext}
\usepackage{multirow}
\usepackage{subcaption}
\usepackage{amssymb}
\usepackage{pifont}

\newcolumntype{P}[1]{>{\centering\arraybackslash}p{#1}}

\theoremstyle{definition}

\theoremstyle{remark}

\theoremstyle{plain}
\newtheorem{theorem}{Theorem}[section]

\theoremstyle{plain}

\theoremstyle{plain}

\newtheorem{proposition}[theorem]{Proposition}

\usepackage{graphicx,wrapfig,lipsum}

\usepackage{amsmath,amsfonts,bm}

\def\eqref#1{(\ref{#1})}

\def\1{\bm{1}}

\DeclareMathAlphabet{\mathsfit}{\encodingdefault}{\sfdefault}{m}{sl}
\SetMathAlphabet{\mathsfit}{bold}{\encodingdefault}{\sfdefault}{bx}{n}

\usepackage{commath}
\usepackage{amssymb}
\usepackage{amsmath,amsfonts,bm}

\newcommand{\hamn}{\{0,1\}^{n}}

\newcommand{\indicator}{\mathbb{I}}
\newcommand{\xk}[1]{\ensuremath x^{\oplus #1}}

\newcommand{\sensi}{\mathcal{S}}
\newcommand{\funcset}{\mathcal{F}}
\newcommand{\vcd}{\mathsf{VCD}}

\newcommand{\pari}{\textsc{Parity}}
\newcommand{\spar}{\textsc{Sparse Parities}}

\newcommand{\parn}[1]{\textsc{Parity}\text{-}#1}
\newcommand{\sparn}[2]{\textsc{Parity}\text{-}(#1,#2)}
\newcommand{\vparn}[2]{\textsc{VarParity}\text{-}(#1,#2)}
\newcommand{\majnk}[2]{\textsc{Maj}\text{-}(#1,#2)}
\newcommand{\varmaj}[2]{\textsc{VarMaj}\text{-}(#1,#2)}
\newcommand{\dictnk}[1]{\textsc{Dict}\text{-}#1}

\newcommand{\kspar}{{k\textsc{-sparse}}}
\newcommand{\parity}{\textsc{Parity}}
\newcommand{\ksparse}{\kspar}
\newcommand{\SPARSEkn}{\textsc{SPARSE}\text{-}(k,n)}
\newcommand{\randkn}{\textsc{Juntas}\text{-}(n,k)}
\newcommand{\rand}[2]{\textsc{Juntas}\text{-}(#1,#2)}

\newcommand{\sparity}{f_\mathsf{parity_k}}
\newcommand{\maj}{f_\mathsf{maj}}
\newcommand{\smaj}{f_\mathsf{maj_k}}

\newcommand{\hatf}{\hat{f}}

\DeclareMathOperator{\Ex}{\mathbb{E}}

\title{Simplicity Bias in Transformers and their Ability \\ to Learn Sparse Boolean Functions}

\author{Satwik Bhattamishra$^\spadesuit$ \quad Arkil Patel$^\diamondsuit$ \quad Varun Kanade$^\spadesuit$ \quad Phil Blunsom$^{\spadesuit \clubsuit}$\\
	\raise3.5ex\hbox{}$^\spadesuit$University of Oxford \quad
	$^\diamondsuit$Mila and McGill University \quad
	$^\clubsuit$Cohere
	\\
	\raise3ex\hbox{}\normalsize{\texttt{\{satwik.bmishra, varun.kanade, phil.blunsom\}@cs.ox.ac.uk}} \quad \normalsize{\texttt{arkil.patel@mila.quebec}}  \\
}

\begin{document}
\maketitle
\begin{abstract}
Despite the widespread success of Transformers on NLP tasks, recent works have found that they struggle to model several formal languages when compared to recurrent models. This raises the question of why Transformers perform well in practice and whether they have any properties that enable them to generalize better than recurrent models. In this work, we conduct an extensive empirical study on Boolean functions to demonstrate the following: (i) Random Transformers are relatively more biased towards functions of low sensitivity. (ii) When trained on Boolean functions, both Transformers and LSTMs prioritize learning functions of low sensitivity, with Transformers ultimately converging to functions of lower sensitivity. (iii) On sparse Boolean functions which have low sensitivity, we find that Transformers generalize near perfectly even in the presence of noisy labels whereas LSTMs overfit and achieve poor generalization accuracy. Overall, our results provide strong quantifiable evidence that suggests differences in the inductive biases of Transformers and recurrent models which may help explain Transformer's effective generalization performance despite relatively limited expressiveness. 
\end{abstract}

\section{Introduction}\label{sec:intro}

Transformers \citep{vaswani2017attention} have supplanted recurrent models across a range of NLP tasks \citep{liu2019roberta, gpt3neurips}. In particular, effective large-scale pretrained models have predominantly been Transformer-based models and have found application in other areas such as computer vision and protein folding. Given the irrefutable importance of understanding these architectures, a significant effort has been devoted to analyze the inner workings of large-scale pretrained Transformers. However, the cause behind the difference in performance between Transformers and recurrent models has largely been unclear. 


A line of work has attempted to understand neural sequence models through the lens of formal language theory. These works have sought to understand the expressive power of these architectures and identify differences in their ability to generalize across various formal languages. A notable result by \citet{hahn-2020-theoretical} showed that Transformers are limited in their ability to express the $\pari$ language\footnote{Computing whether a bit string has odd or even number of ones.} while it is well known that small-sized RNNs can express such languages. Across empirical studies, Transformers have been found to perform worse or comparably to LSTMs in almost all formal languages previously considered in the literature \citep{bhattamishra-etal-2020-ability, deepminddeletang2022neural, chiang-cholak-2022-overcoming}. In particular, Transformers have been shown to struggle with the $\pari$ language and certain other regular languages. This leads to a natural question: Why do Transformers perform so well in practice if they are arguably less expressive and perform worse than LSTMs across certain formal languages? 

\begin{figure*}[t]%
	\centering
	\subfloat{{\includegraphics[scale = 0.27, trim=0 0 5 5, clip]{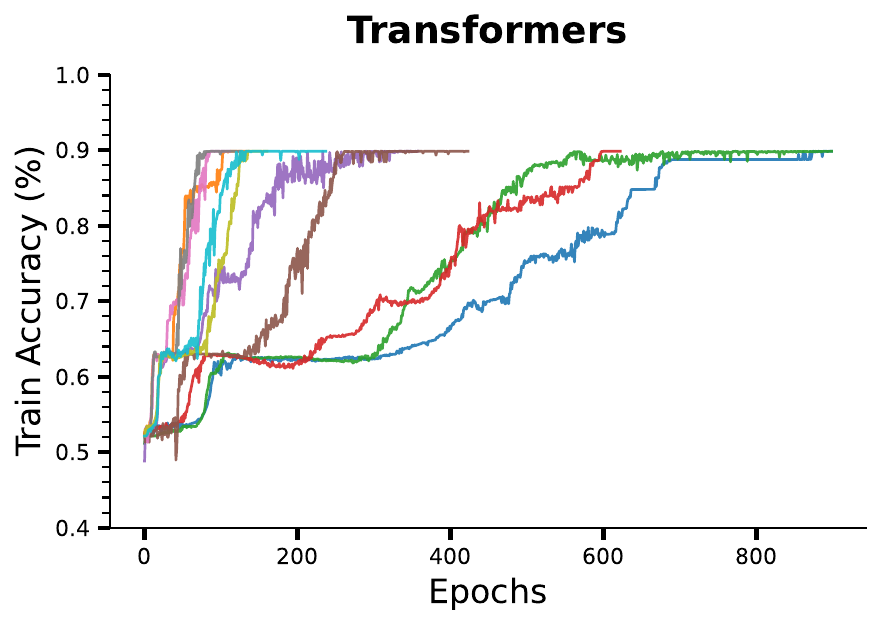}  }}%
	\subfloat{{\includegraphics[scale = 0.27, trim=0 0 5 5, clip]{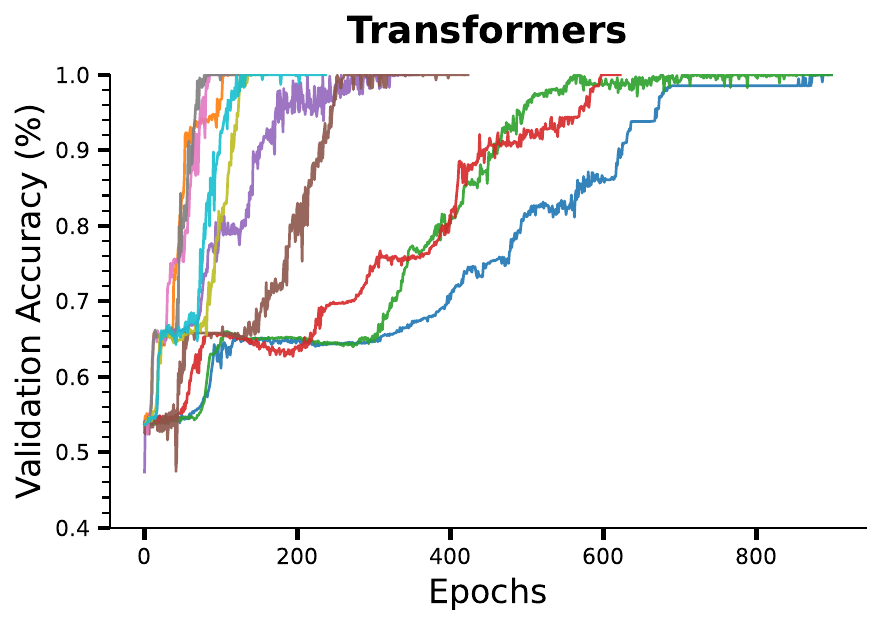}   }}
	\subfloat{{\includegraphics[scale = 0.27, trim=0 0 5 5, clip]{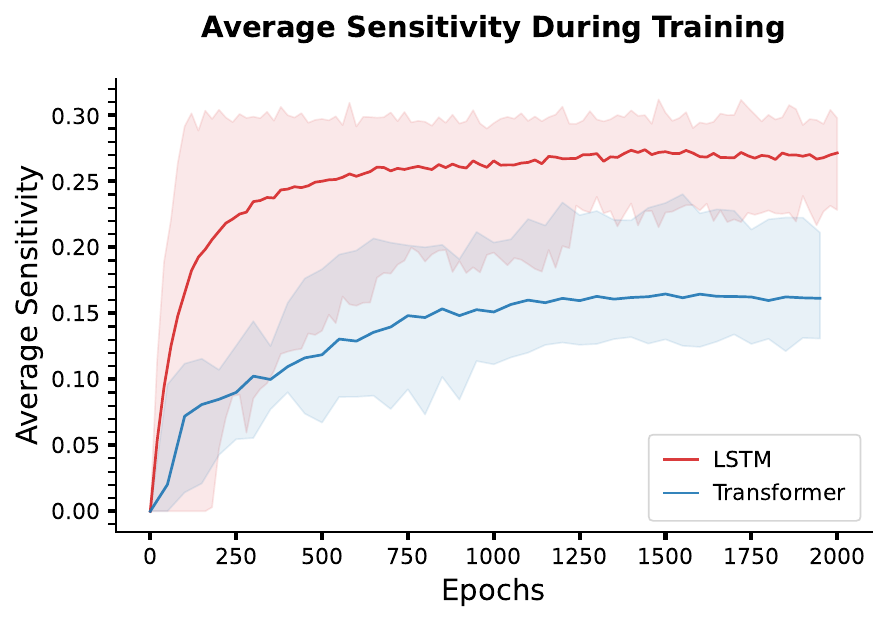}  }}
    \medskip
    
    \subfloat{{\includegraphics[scale = 0.27, trim=0 0 5 5, clip]{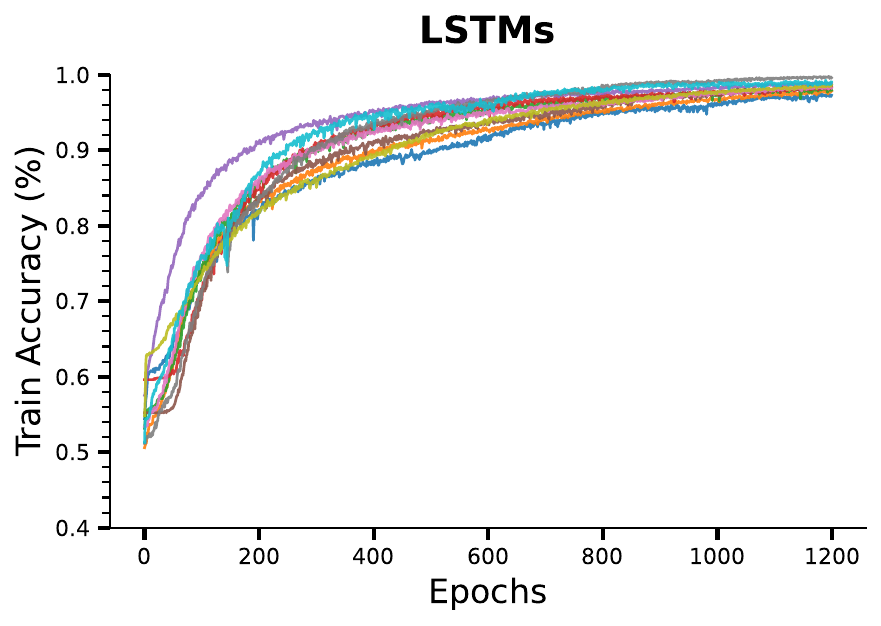}  }}%
	\subfloat{{\includegraphics[scale = 0.27, trim=0 0 5 5, clip]{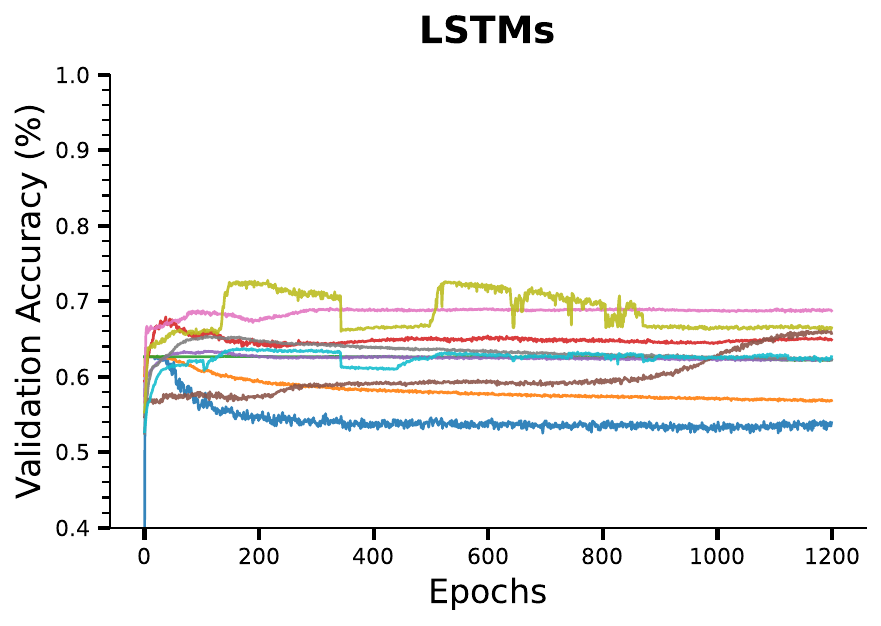}  }}
	\subfloat{{\includegraphics[scale = 0.31, trim=0 0 5 5, clip]{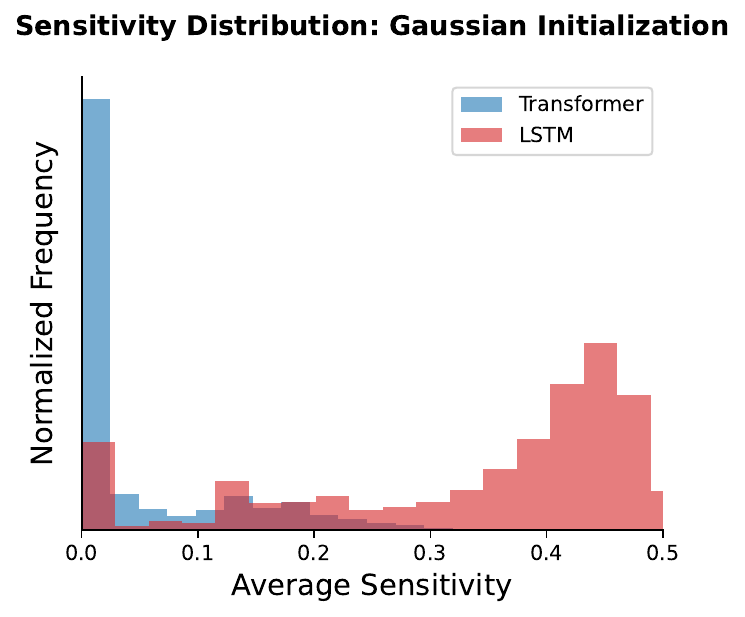} }}
	
	\caption{ Overview of the main findings of the paper. Training and validation curves for Transformers (top left) and LSTMs (bottom left) on 10 different $\kspar$ functions in the presence of 10\% noise. Sensitivity of Transformers and LSTMs during training on random Boolean functions (top right). Distribution of sensitivity of functions represented by random models with weights initialized according to normal distribution ($\mu$=0, $\sigma$=10)  (bottom right).}%
	\label{fig:intro}%
\end{figure*}

Although recurrent models such as LSTMs have been shown to perform better on formal languages such as $\pari$, we find that they struggle to generalize well on several sparse Boolean functions such as $\spar$. We find a clear contrast between the generalization abilities of Transformers and LSTMs on various $\kspar$ Boolean functions which have low sensitivity. Additionally, through extensive empirical analysis, we provide strong evidence to suggest differences in the bias towards low complexity functions between Transformers and recurrent models. Based on our results, we hypothesize that one of the reasons behind Transformer's practical effectiveness could be that they are more biased towards simple functions in comparison to recurrent models which may lead to better generalization.


  In particular, we focus on a complexity measure called sensitivity \citep{kahn1989influence_sensitivity}, which measures how likely it is that a function value changes due to a `small' change in input. Sensitivity is related to several other complexity measures; functions with low sensitivity have low Kolmogorov complexity,\footnote{The converse is not always true.} simpler Fourier spectra, and can be represented by decision trees of small depths. The relationship between sensitivity and generalization has also been previously studied in the literature \citep{novak2018sensitivity,franco2006generalization}.\footnote{A more thorough discussion is presented in Appendix \ref{subsec:why_sensi}} While measures such as Kolmogorov complexity are uncomputable, sensitivity can be tractably estimated and extensions of sensitivity can be used to estimate the complexity of functions in more realistic NLP tasks \citep{hahn-etal-2021-sensitivity}.


\textbf{Our Contributions.}  We investigate the bias in (a) \emph{parameter space} by analyzing randomly initialized models, and (b) \emph{learning procedure} by examining the sensitivity of models during the training process. Motivated by our findings indicating differences between the biases of Transformers and LSTMs, we evaluate their performance on functions of low sensitivity.

 \textbf{(i)} We demonstrate that random Transformers are significantly more likely to represent functions of lower sensitivity than recurrent models when the weights are sampled uniformly or according to Normal distribution (see Figure \ref{fig:intro}, bottom right). When the weights are initialized following practical strategies (such as Xavier normal), then both architectures are likely to have low sensitivity with Transformers having relatively lower sensitivity.


 \textbf{(ii)} We show that both Transformers and LSTMs learn functions of increasing sensitivity when trained on a set of Boolean functions as well as practical datasets such as sentiment classification (see Figure \ref{fig:intro}, top right). For Boolean functions, Transformers converge to functions of lower sensitivity in comparison to LSTMs when both models achieve near-zero training error.

 \textbf{(iii)} On various $\kspar$ Boolean functions, we find that Transformers generalize near-perfectly even in the presence of noise in the training data whereas LSTMs severely overfit and obtain poor generalization performance (see Figure \ref{fig:intro}, left).

\textbf{Auxiliary Results.} Although not the primary focus of the paper, we explore relations between sensitivity and generalization in Appendix \ref{app:sensi_gen}. In particular, we show how sensitivity can be used as a capacity measure to derive generalization bounds. Additionally, we explore the correlation between sensitivity and generalization gap for LSTMs and Transformer-based models on sentiment classification tasks. We also 
conduct experiments with three other complexity measures in Appendix \ref{appsub:complexity}.

\section{Related Work}\label{sec:rel_work}

\textbf{Random Neural Networks.} One approach to explaining deep learning's unexpected generalization performance has been to study the inductive biases of random neural networks. Several prior works have shown theoretically \citep{de-palma19-simple} and empirically \citep{valle-perez2018deep} that random untrained feedforward networks are biased towards `simple' functions. \citet{valle-perez2018deep} showed that considering the distribution over functions generated via random neural networks as a prior leads to better PAC-Bayesian generalization bounds than traditional ones. Several works \citep{mingard2019neural,randwilson2020bayesian,randlee2017deep} have argued using heuristic methods that the inductive biases in random neural networks can be used to understand the properties of trained networks. Additionally, there is empirical and theoretical evidence \citep{closeinitoymak2019overparameterized} that neural networks trained with SGD usually converge close to the initialization point. Hence, understanding the properties of random neural networks is imperative to understand their generalization abilities.\footnote{There is an enormous literature on this topic; refer to \citet{bayesiansamplerSgd} for more references and discussion.} In Section \ref{subsec:randsensi_exp}, we study the complexities of random Transformers and recurrent models and investigate the differences between them. 


\noindent \textbf{Formal Languages and Sequence Models.} In the past few years, a strand of work\footnote{See Appendix \ref{app:relwork} for a more comprehensive version.} primarily in the NLP community has attempted to understand neural sequence models' capabilities and inner workings by analyzing them on formal languages, e.g. \citep{suzgun2018evaluating,sennhauser-berwick-2018-evaluating}. Given the recent success of Transformers, several works have sought to investigate them via the lens of formal languages. \citet{hahn-2020-theoretical} theoretically showed the limitations of Transformers in recognizing languages like Parity and Dyck-2. While Transformers are expressive enough to represent the $\pari$ language for bounded lengths \citep{chiang-cholak-2022-overcoming}, multiple works have observed that they struggle to generalize well on Parity and other regular languages when tested empirically \citep{bhattamishra-etal-2020-ability, chiang-cholak-2022-overcoming, deepminddeletang2022neural}. In contrast to this, we show that when evaluated on some simpler variants of these formal languages, Transformers generalize near perfectly whereas LSTMs achieve poor generalization performance.

\section{Background and Preliminaries}\label{sec:background}

\subsection{Sensitivity of Boolean Functions}

We will work with a complexity measure called Boolean Sensitivity which has been widely studied in computational complexity \citep{kahn1989influence_sensitivity,sensi_other_2014tighter}. Sensitivity can be seen as a discrete analog \citep{gopalan2016smooth} of the `smoothness' of a continuous function which measures how gradually a function changes locally. For Boolean functions defined over the Hamming cube, sensitivity captures how many neighbours of a particular input have different outputs. Formally, the sensitivity of a Boolean function $f:\{0,1\}^n \rightarrow \{\pm1\}$ at input $x \in \hamn$ is defined as 

\begin{equation}\label{eq:sensi_input}
    s(f,x) = \sum_{i=1}^{n} \indicator[f(x) \neq f(\xk{i})],     
\end{equation}
where $\indicator$ denotes the indicator function and $\xk{i} = (x_1,\ldots,x_{i-1},1-x_i,x_{i+1},\ldots,x_n)$ is the same as $x$ at every coordinate or bit except the $i$-th one. The maximum sensitivity of a function $f$ is defined as $ms(f) = \max_{x\in \hamn} s(f,x)$. The average sensitivity (also referred to as total influence) of a Boolean function measures the average of the sensitivity of the function across all inputs $x\in \hamn$ and is defined as 

\begin{equation}\label{eq:avg_sensi}
    s(f) = \Ex_x[s(f,x)]= \frac{1}{2^n} \sum_{x\in \hamn} s(f,x).     
\end{equation}

See that $0 \leq s(f)\leq ms(f) \leq n$. To compare across inputs of different lengths, in our experiments we will normalize the average sensitivity across length $\sensi(f) = \frac{1}{n}s(f)$ which can also be interpreted as,

\begin{equation}\label{eq:sensi}
    \sensi(f) = \Pr_{x \sim  \hamn, i \sim [n]}[f(x) \neq f(\xk{i})]
\end{equation}

\noindent where $[n]= \{1,\ldots, n\}$ and the sampling is over uniform distribution over the domains. 

\textbf{Parity.} The Parity function over $\hamn$ is defined as $\parity(x) := (-1)^{\sum_{i=1}^n x_i}$. For any input $x \in \hamn$, the function $\parity$ has value $+1$ if the number of ones in the input is even and has value $-1$ otherwise. The sensitivity of the Parity function is the maximum among all functions since changing any bit of any input changes the function value. Hence, for $\parity$ over $\hamn$, $s(\parity) = n$ and $\sensi(\parity) = 1$. 

\textbf{Sparse Boolean functions.} Another class of functions are the $\ksparse$ functions (also referred to as $k$-juntas) where the function value depends on at most $k$ coordinates of the input. More formally, a function $f:\hamn \rightarrow \{\pm 1\}$ is $\ksparse$ if there exist indices $1 \leq i_1 < i_2 < \ldots <i_k \leq n$ and a function $g: \{0,1\}^k \rightarrow \{\pm 1\}$, such that for every $x \in \hamn$, $f(x_1, x_2, \ldots, x_n) = g(x_{i_1}, x_{i_2}, \ldots, x_{i_k})$. Let $\SPARSEkn$ be the class of $\ksparse$ functions on inputs of length $n$ that depend on at most $k$ bits. It is easy to see that, for any $f \in \SPARSEkn$, the average sensitivity $s(f) \leq k$ (and hence $\sensi(f) \leq \frac{k}{n}$). 

When $k \ll n$, $\SPARSEkn$ can be seen as a subclass of all Boolean functions with low average sensitivity. Other functions with low average sensitivity can also be approximated with $\ksparse$ functions using Friedgut’s Junta Theorem (\citet{bool_analysis}, Page 269). The maximum average sensitivity  $s(f) = k$ is attained by $\spar$ denoted $\sparity$ which is the Parity over a subset of $k$ coordinates. A sparse parity function $\sparity$ over $S \subseteq [n]$, s.t. $|S|=k$ is $+1$ if the number of ones in the coordinates $S$ is odd and $-1$ otherwise. Other Boolean functions such as sparse majority can be defined similarly. The majority function $\maj$ over $\hamn$ is $+1$ if the number of ones in the input is greater than the number of zeros and is $-1$ otherwise. Similarly, the sparse majority function $\smaj$ is the majority function over coordinates $S \subseteq [n]$, s.t. $|S|=k$. Parities (and Sparse Parities) are an important class of Boolean functions since any Boolean function can be represented as a linear combination of a set of Parity functions.

\section{Sensitivity Experiments}\label{sec:sensi_exp}

\begin{figure}[t]%
	\centering
	\subfloat{{\includegraphics[scale = 0.3, trim=0 0 5 5, clip]{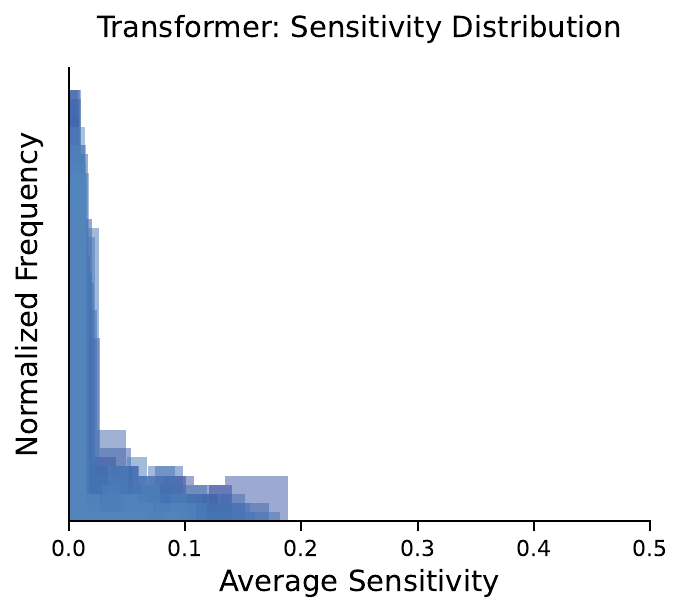} \label{fig:uni_san} }}%
	\subfloat{{\includegraphics[scale = 0.3, trim=0 0 5 5, clip]{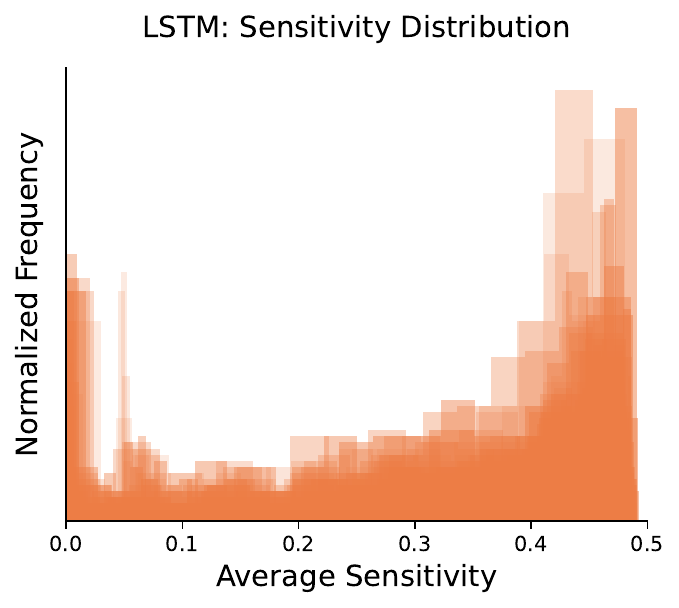}  \label{fig:uni_rnn} }}
    \medskip
    
    \subfloat{{\includegraphics[scale = 0.3, trim=0 0 5 5, clip]{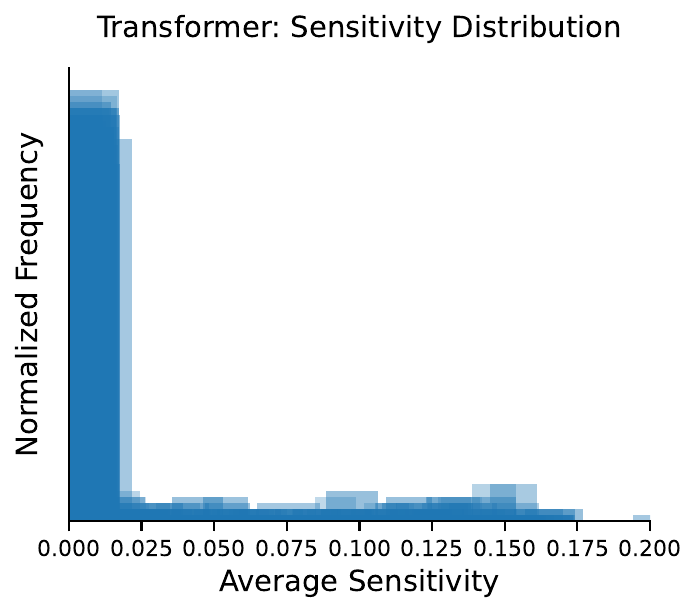} \label{fig:xav_san} }}%
	\subfloat{{\includegraphics[scale = 0.3, trim=0 0 5 5, clip]{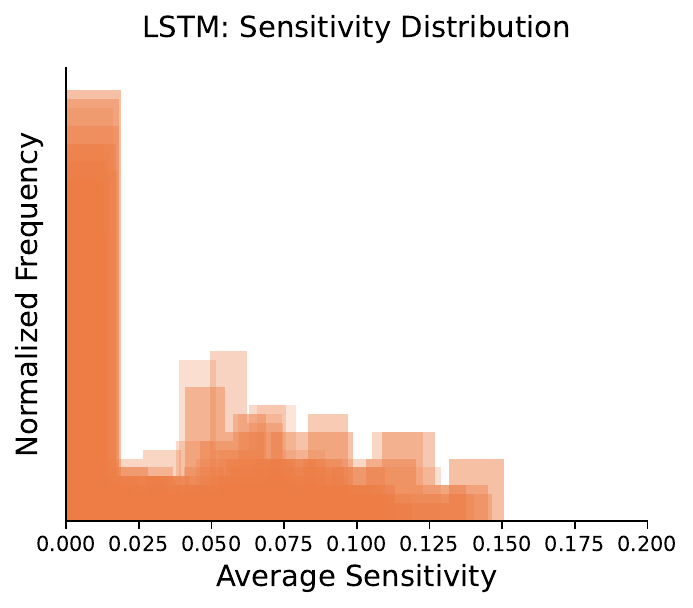} \label{fig:xav_rnn} }}
	
	\caption{Distribution of sensitivity of different randomly initialized Transformers and LSTMs. \textit{Top row:} Transformers (left) and LSTMs (right) with uniformly sampled weights across various hyperparameters. \textit{Bottom row:} Analogous to top row but with Xavier normal initialization. Refer to Section \ref{subsec:randsensi_exp} for details.}%
	\label{fig:sensi_uni}%
\end{figure}

In this section, we conduct various experiments to investigate the differences in the bias of Transformers and RNNs towards functions of low sensitivity.\footnote{We have made our source code available at \href{https://github.com/satwik77/Transformer-Simplicity}{https://github.com/satwik77/Transformer-Simplicity}.} From here onward, whenever sensitivity is mentioned, we will refer to the length normalized version of average sensitivity $\sensi$ defined in Eq. (\ref{eq:sensi}). The first part of this section deals with analyzing the sensitivity of random Transformers and RNNs while the second part investigates the sensitivity of models trained to fit random Boolean functions.


\subsection{Sensitivity of Randomly Initialized Models}\label{subsec:randsensi_exp}

We seek to understand the landscape of the complexity of functions in the parameter space of Transformers and RNNs. Let us assume that the parameter space $\Theta$ of our models is bounded, i.e. all the parameters (weights) take some value within some bounded range $[-B, B]$. A particular realization of the parameters with values in $[-B, B]$ leads to the model being a function from $\hamn \rightarrow \{0,1\}$. We begin with a simple question: Out of all the parameterizations in the parameter space of Transformers (or RNNs), if we select one uniformly at random, then how likely is it to have low sensitivity?

\textbf{Setup.} In all our experiments, we consider binary classifiers with Transformers and RNN-based architectures. By Transformer, we refer to the encoder-only version of the original Transformer architecture \citep{vaswani2017attention} as used in models such as BERT \citep{devlin-etal-2019-bert}. The model takes a sequence of tokens along with a $\textrm{[CLF]}$ token as input. The final classification is done based on the output vector of the $\textrm{[CLF]}$ token. For recurrent models, we consider LSTMs \citep{lstm_hochreiter1997long}, GRUs, and RNNs with \texttt{tanh} activation. Most of the results in the main paper pertaining to recurrent models are based on experiments with LSTMs and we discuss when the behaviour is different for other recurrent models.

In our experiments, we explore four strategies to sample random networks: Uniform, Gaussian, Xavier uniform, and Xavier normal initialization. In uniform sampling, each parameter (weights and biases) is assigned a value by uniformly sampling in $[-10, 10]$. Similarly, for Gaussian initialization, each parameter is assigned by sampling from $\mathcal{N}(0, \sigma^2)$ where we set sigma as $10$. Xavier normal \citep{xavierglorot2010understanding} initialization is the one that is more commonly used in practice to train these models. All the weights are initialized with $\mathcal{N}(0, \sigma^2)$ where the standard deviation $\sigma = d^{-1/2}$ where $d$ is the number of hidden units. All the input embedding vectors and positional embedding vectors are initialized with $\mathcal{N}(0, 1)$ which is the default scheme in PyTorch \citep{pytorch}. For input lengths greater than 10, we estimate the sensitivity of each model by computing the average over a sampled set of bit strings. We sample 10k bit strings and compute the average sensitivity across the samples. For each hyperparameter configuration, we sample 75-1000 different models to estimate their sensitivity depending on the computational costs associated with it. For most of the results reported in the main paper, we consider bit strings of length $20$. But we also experiment with lengths $\in \{5, 7, 10, 15, 20, 50, 100, 200\}$.

\begin{figure}[t]%
	\centering
	\subfloat{{\includegraphics[scale = 0.3, trim=0 0 5 5, clip]{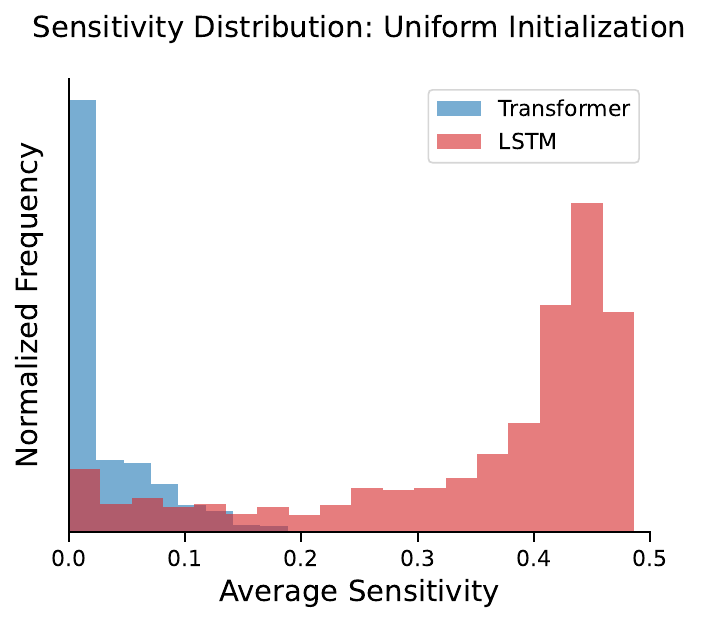} \label{fig:uni_over} }}
	\subfloat{{\includegraphics[scale = 0.3, trim=0 0 5 5, clip]{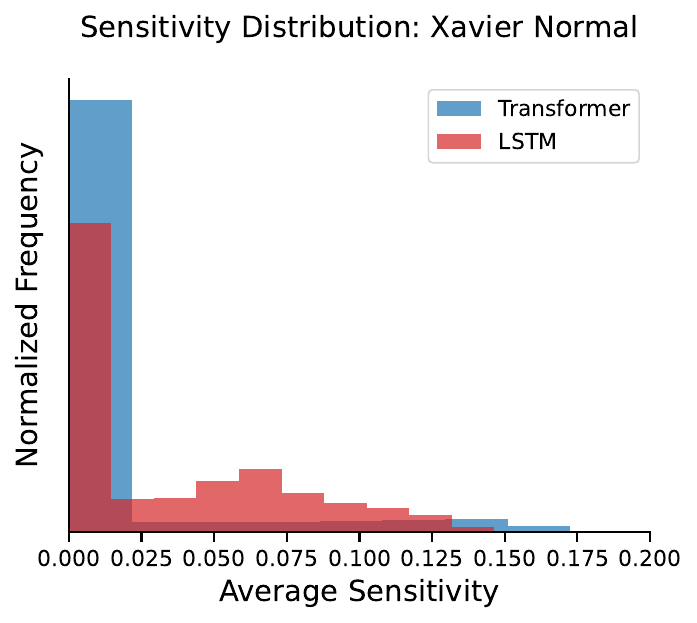} \label{fig:xav_over} }}

	
	\caption{ \label{fig:sensi_fix} Distribution of sensitivity of randomly initialized Transformers and LSTMs for a fixed hyperparameter (\texttt{layers}=2, \texttt{width}=256).}%
	
\end{figure}

\textbf{Results.} Figure \ref{fig:sensi_uni} (upper row) and Figure \ref{fig:sensi_fix} (left) shows the distribution of sensitivity for uniformly initialized Transformers and LSTMs. The distribution for Transformers is heavily skewed towards functions of very low sensitivity in comparison to LSTMs. The pattern holds across Gaussian initialization as well (see Figure \ref{fig:intro}, bottom right). For initialization strategies used in practice such as Xavier normal and Xavier uniform, we find that both Transformers and LSTMs have low sensitivity (see Figure \ref{fig:sensi_uni}, lower row and Figure \ref{fig:sensi_fix}, right) with Transformers having relatively lower average sensitivity. Refer to Section \ref{appsub:addsensi} in the Appendix for results with Xavier uniform initialization. 

Although we primarily discuss results with sensitivity in the main paper, similar experiments with other complexity measures are presented in Appendix \ref{appsub:complexity}. Further experiments exploring the change in distribution across the number of layers, width, and lengths for both architectures are presented in Appendix \ref{appsub:addsensi}.

\begin{figure}[t]
\centering
   \includegraphics[scale=0.4]{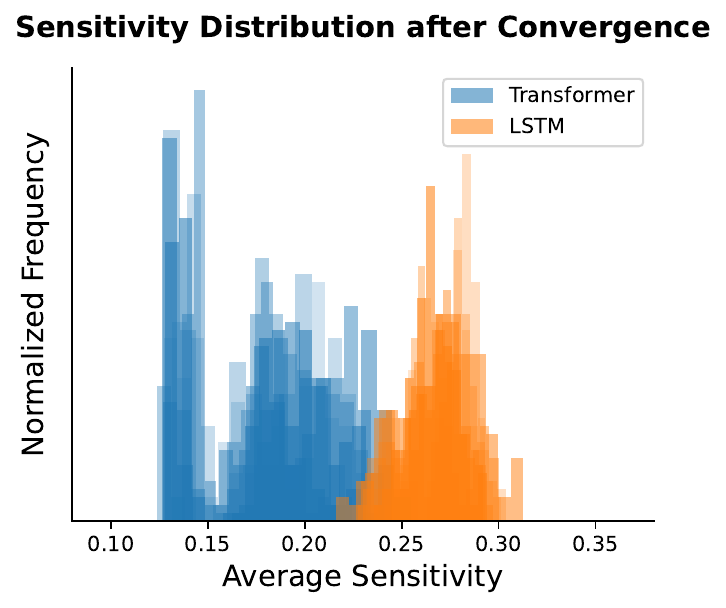}
	\caption{\label{fig:rnnsan_conv} Distribution of sensitivity of Transformers and LSTMs trained on Boolean strings with random labels. Refer to Section \ref{subsec:randbool_exp} for details.}
	
\end{figure}

\textbf{Discussion.} These results imply that lower sensitivity functions are \textit{over-represented} in the parameter space of Transformers. If every Boolean function $f:\hamn\rightarrow\{0,1\}$ would have had equal representation in the parameter space of the model, then the distribution would have concentrated around $1/2$. A learning algorithm based on a random search over the parameter space is more likely to encounter functions of low sensitivity. Note that, while recurrent models have higher sensitivity than Transformers, they are still lower than randomly sampling a Boolean function.


\textbf{Why randomly initialized models?}  Each randomly initialized Transformer or RNN when restricted to domain $\hamn$ represents one of the $2^{2^{n}}$ Boolean functions $f$ in $\funcset$. The distribution over $\funcset$ induced by randomly initialized models can be seen as their prior $P(f)$. Given a set of (training) examples $S = \{(x_1, y_1), \ldots, (x_m, y_m)\}$, let $P_B(f|S)$ denote the probability of sampling $f$ conditioned on the event that the sampled function is consistent with $S$ (matches all the input-output mappings in $S$). We can apply Bayes' rule to calculate the posterior $P_B(f|S) = P(S|f)P(f)/P(S)$ using the prior $P(f)$, the likelihood $P(S|f)$, and the marginal likelihood $P(S)$. Since we condition on $f$ being consistent with $S$ (zero training error), the likelihood $P(S|f) = 1$ if $\forall x_i \in S$, $f(x_i) = y_i$ and $0$ otherwise. Let $U(S)$ denote the set of all functions $f \in \funcset$ which are consistent with $S$. Note that, given a fixed training set $S$, since $P(S)$ is constant, the probability over the choice of $f \in U(S)$ ultimately depends on the prior $P(f)$. In practice, we do not fit the training set by sampling models to find one that is consistent with the training set. However, recent work \cite{bayesiansamplerSgd} has shown that $P_B(f|S) \approx P_{SGD}(f|S)$ across a range of neural architectures and data sets. The SGD-based posterior $P_{SGD}(f|S)$ denotes the probability that a neural network converges on function $f$ when trained to fit $S$. Hence, our results suggest that for Transformers, $P_B(f|S)$ would be concentrated on low-sensitivity functions and consequently,  $P_{SGD}(f|S)$ could be biased towards low-sensitivity functions as well.

\subsection{Models learn functions of increasing sensitivity}\label{subsec:randbool_exp}

In this section, we investigate the sensitivity of functions learned during the training process when Transformers and LSTMs are trained to fit datasets of Boolean strings with random labels.

\textbf{Setup.} We create datasets of size $1$k each by uniformly sampling bit strings of length $40$. The label for each input string is assigned randomly ($+1$ or $-1$ with probability $1/2$). All the weights of the models are initialized with Xavier normal initialization and the biases are initialized with zero vectors. We consider Transformers and LSTMs across various hyperparameter configurations with a similar number of parameters. We train the models until they reach zero training error and estimate the sensitivity of the models at every epoch. We conduct the experiments over $20$ different datasets with $100$ runs for Transformers and LSTMs each.

\textbf{Sensitivity during training.} We find that both Transformers and LSTMs gradually learn functions of increasing sensitivity with Transformers converging to functions of much lower sensitivity than LSTMs (refer to Figure \ref{fig:intro}, top right). We observe similar behavior when the models are trained on various sparse Boolean functions including sparse parities. Even though sensitivity is defined over Boolean functions, we explore a few natural extensions to estimate the sensitivity of models trained on real datasets such as sentiment classification. On two sentiment classification datasets, namely SST and IMDB, we found similar observations where both Transformers and LSTMs seem to incrementally learn functions of increasing sensitivity. See Appendix \ref{app:sensi_real} for more details.

\begin{figure}[t]%
	\centering
	\subfloat{{\includegraphics[scale = 0.255, trim=0 0 5 5, clip]{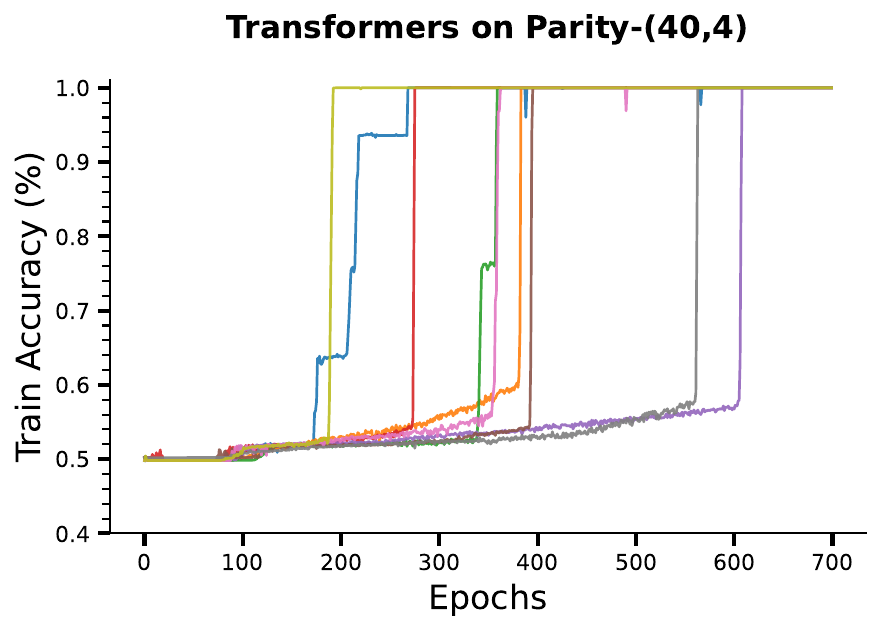} \label{fig:san_sparity_tracc} }}%
	\subfloat{{\includegraphics[scale = 0.255, trim=0 0 5 5, clip]{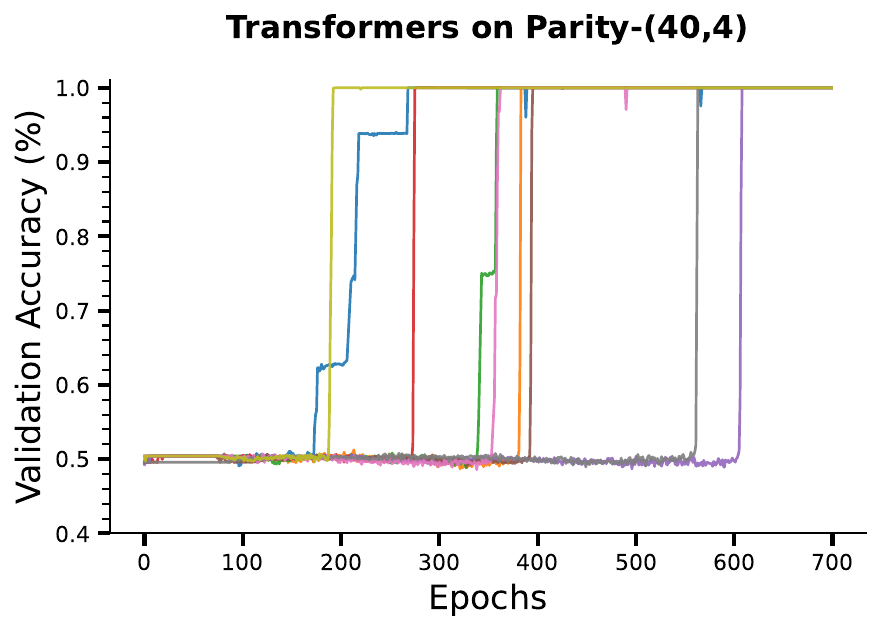} \label{fig:san_sparity_valacc} }}
    \medskip
    
    \subfloat{{\includegraphics[scale = 0.255, trim=0 0 5 5, clip]{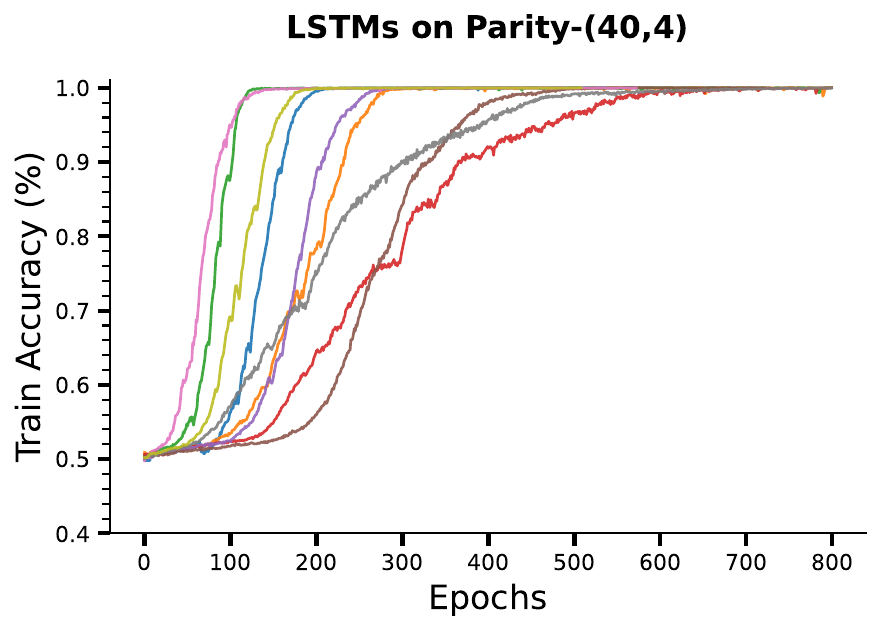} \label{fig:rnn_sparity_tracc} }}%
	\subfloat{{\includegraphics[scale = 0.255, trim=0 0 5 5, clip]{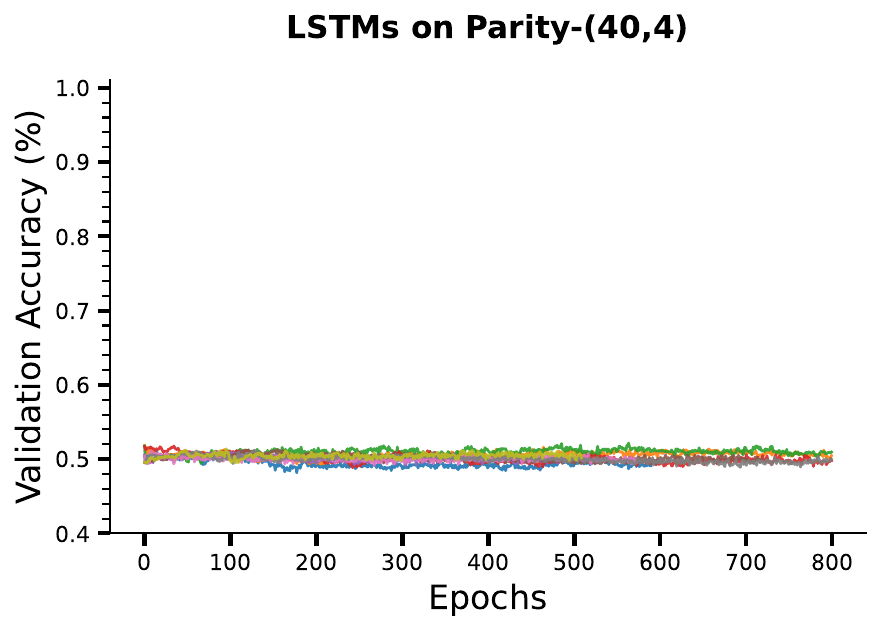} \label{fig:rnn_sparity_valacc} }}
	
	\caption{Training and validation curves for Transformers and LSTMs trained on $\spar$ of length $n=40$ and $k=4$.}%
	\label{fig:sparity_results}%
\end{figure}

\textbf{Discussion.} Even if sequence models such as Transformers or LSTMs are capable of representing arbitrary functions, our results suggest that they prioritize learning simpler patterns first. These results echo prior observations that indicate feedforward-like neural networks trained with SGD learn functions of increasing complexity \citep{sgd_complexity, arpit_memorization_icml}. \citet{spectral-rahaman19a} find that ReLU neural networks learn functions of lower frequency modes first. Functions with lower average sensitivity also have a lower frequency and hence these observations are closely connected. More importantly, average sensitivity can be naturally extended to real data which allows us to empirically explore this for text data.

\textbf{Sensitivity upon convergence.} For Transformers and LSTMs trained until $0\%$ training error, we estimate the sensitivity of functions learned by the models. We create 15 datasets and for each dataset, we compute the sensitivity of $100$ trained models. The combined distribution of the sensitivity of the models across all datasets is shown in Figure \ref{fig:rnnsan_conv}. We observe that Transformers consistently learn functions of lower sensitivity in comparison to LSTMs. This supports our hypothesis that for Transformers the parameter search via algorithms such as Adam is more likely to find functions of lower sensitivity that fit the training set as opposed to LSTMs.

\section{Experiments on Sparse Boolean Functions}\label{sec:bool_exp}

Our results in the previous section indicate that relative to LSTMs, random Transformers are biased towards low-sensitivity functions and Transformers are biased towards learning Boolean functions of low sensitivity. Motivated by this difference in bias, we conduct experiments geared towards answering the following question: Is there any difference between the ability of Transformers and LSTMs to learn sparse Boolean functions which have low sensitivity?

\subsection{Setup.}\label{subsec:setup}

\textbf{Boolean Functions.} We focus on $\ksparse$ Boolean functions which have low sensitivity when $k \ll n$ (refer to Section \ref{sec:background} for definition). We first consider certain Boolean functions which are widely studied in the analysis of Boolean functions. The first one is $\spar$ which can be interpreted as the $\ksparse$ variation of standard parity. We denote an instance of $\spar$ as $\sparn{n}{k}$ where $n$ denotes the length of the input string and $k$ denotes the number of relevant bits. We denote an instance of standard $\pari$ as $\parn{n}$ where $n$ denotes the length of the input string and the output is computed based on the number of ones in all indices. Learning $\sparn{n}{k}$ with gradient-based methods has well-known hardness results $-$ requiring at least $n^{\Omega(k)}$ computational steps to find the correct target function \citep{kearns1998efficient}. The other two Boolean functions we consider are sparse majorities (denoted by $\majnk{n}{k}$) and the dictator function (denoted by $\dictnk{n}$). The output of the dictator function depends only on a single input bit, making it arguably one of the simplest Boolean functions with very low sensitivity. In $\majnk{n}{k}$, the output for a string of length $n$ is determined by whether the number of ones is greater than the number of zeros in the $k$ relevant indices.

The second set of Boolean functions we consider is random $\ksparse$ functions (denoted by $\randkn$). For each instance of $\randkn$, the function is determined by randomly choosing $k$ indices and assigning labels to each of the $2^k$ distinct inputs randomly.\footnote{Note that this is separate from the random Boolean functions in Section \ref{subsec:randbool_exp} where labels of unseen data is also random. In this scenario, the labels of the unseen inputs can be determined by learning the correct function from the training set.}

\begin{figure}[t]%
	\centering
	\subfloat{{\includegraphics[scale = 0.26, trim=8 0 12 5, clip]{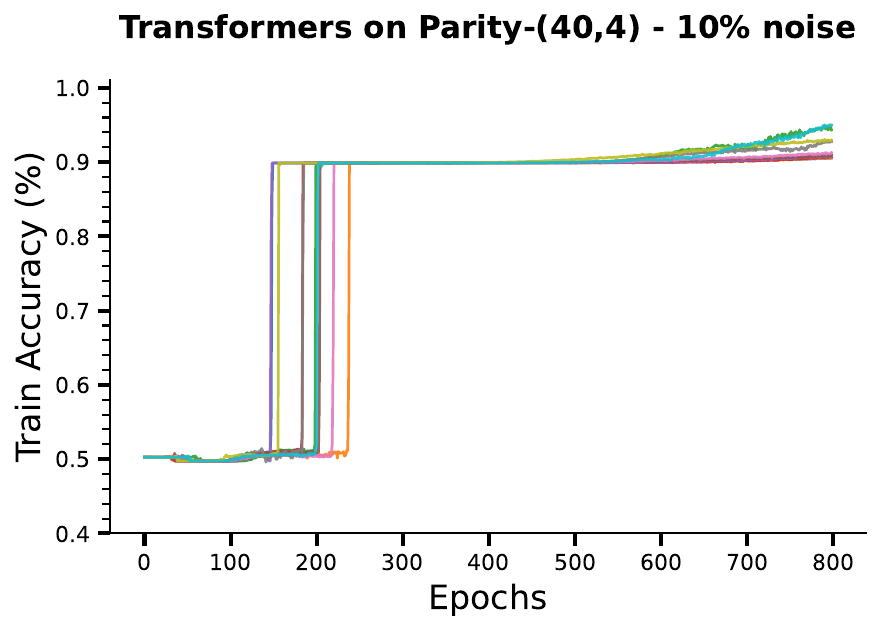}  }}
	\subfloat{{\includegraphics[scale = 0.26, trim=8 0 12 5, clip]{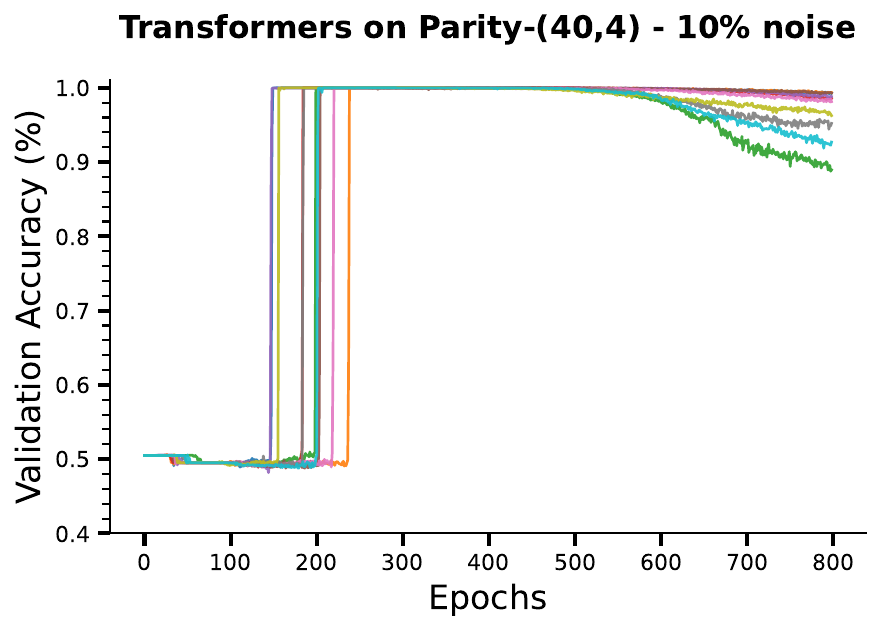}  }}

	
	\caption{ \label{fig:san_sparity_n} Training and validation curves for Transformers trained on $\sparn{40}{4}$ with $10\%$ noise in the training data. See Section \ref{sec:bool_exp} for details.}%
	
\end{figure}

\textbf{Noisy Labels.} We also conduct experiments to examine the ability of the models to learn in the presence of noise. In these experiments, labels of training data are flipped with a certain probability $\eta$. Thus, about $1-\eta$ fraction of the training data is clean and $\eta$ fraction of the training data has incorrect labels. The validation set is clean without any modifications. The goal is to investigate whether a model is robust to noise during the training process.

\textbf{Training Details.} The training and validation sets are created by uniformly sampling bit strings over $\{0, 1\}^{n}$. In our experiments, we consider Transformers with 1-6 layers, 4-8 heads and width (usually referred to as \texttt{d\_model}) within 8-128. We consider Transformers with both learnable and absolute positional encodings. For LSTMs, we consider up to 6 layers and widths (also referred to as \texttt{hidden\_size}) within 8-256. The size of the token embeddings is kept the same as the width. We also consider the presence of learnable positional embeddings as a hyperparameter. We use batch sizes of 100 and 500 in all our experiments and tune across learning rates $\in \{\text{1e-1, 5e-2,} \ldots\text{, 1e-6}\} $. For each dataset, we extensively tune the models across various hyperparameters, details of which are provided in Appendix \ref{app:implementation}.

\subsection{Experiments}\label{subsec:boolexp_results}

\textbf{Parities.} For $\sparn{40}{4}$ and $\parn{40}$, we create $5$ different datasets and report the results based on the maximum accuracy achieved on unseen test data. The train set consists of 30k samples and the validation sets contain 10k samples.

We observe a stark contrast between the performance of Transformers and LSTMs on different forms of parity tasks. We find that Transformers struggle to fit and generalize on $\parn{40}$ while LSTMs easily (across a range of hyperparameters) generalize well on them. On the other hand, perhaps surprisingly, on $\sparn{40}{4}$, we find that while Transformers generalize well, LSTMs severely overfit and achieve poor validation accuracy. Although LSTMs achieve $100\%$ training accuracy over the training data, their validation accuracy does not move far beyond the chance level ($50\%$). Figure \ref{fig:sparity_results} depicts the training and validation accuracy curves for Transformers and LSTMs on $\sparn{40}{4}$ task. We find similar behaviour for LSTMs even with learnable positional embeddings.

\begin{figure}[t]
	\centering
	\includegraphics[scale = 0.3, trim=5 10 5 5, clip]{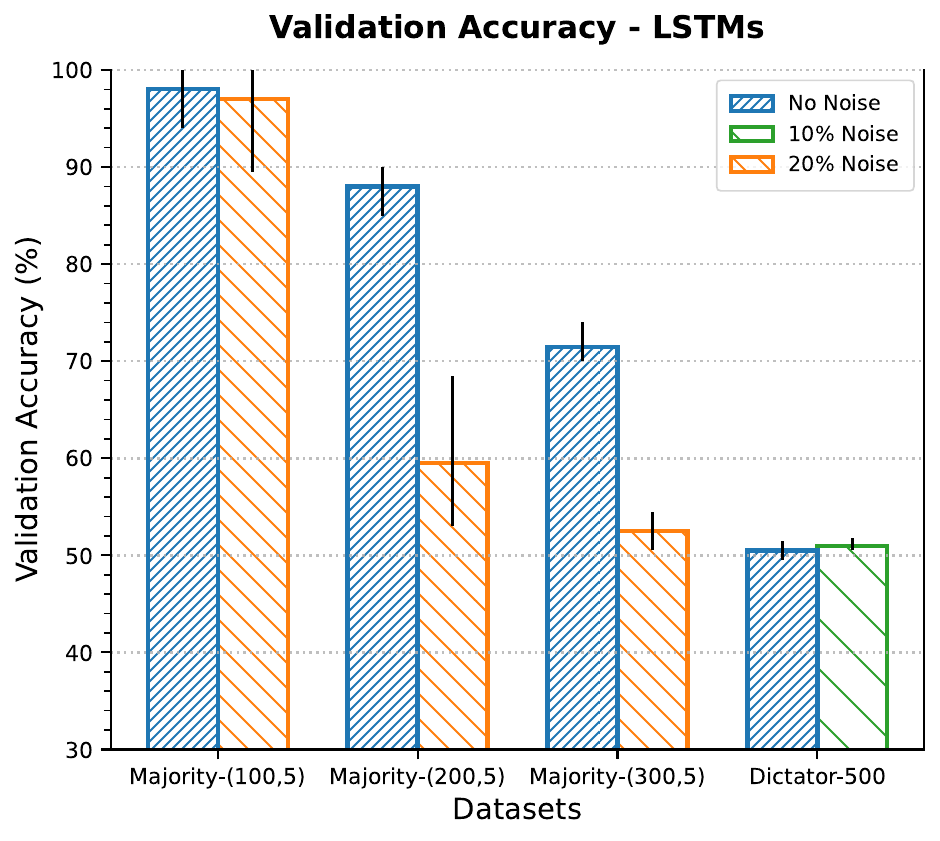}
	\caption{Validation accuracies of LSTMs on sparse Majority and Dictator datasets averaged over 5 runs for best performing hyperparameter.}
	\label{fig:rnn_majdict}
\end{figure}

\textbf{Robustness to noise.} On $\spar$ datasets, we find that Transformers are surprisingly robust to noise. When the training data contains 5\%-20\% noise ($\eta$), Transformers achieve perfect generalization accuracy with training accuracy converging at $1-\eta$. In some cases, after training for a large number of iterations post-convergence, Transformers begin to overfit on the noise. This observation echoes a similar finding in \citet{ruder-2022-memorisation} where they observed such behaviour while finetuning large pretrained models for sequence tagging tasks in the presence of noise. The training and validation accuracy curves are provided in Figure \ref{fig:san_sparity_n}. The behaviour of recurrent models is the same as in the previous scenario with clean data: they overfit on the training data while achieving chance level validation accuracy. Additional results on $\sparn{n}{k}$ across different dataset sizes, task variations, as well as exploring phenomena such as phase transitions and grokking are provided in Appendix \ref{app:parities}.

We observe this pattern across other sparse Boolean functions such as sparse majority and dictator functions as well. For sparse majority datasets $\majnk{n}{5}$, we consider lengths $n \in \{50, 75, 100, 200\}$  and for dictator functions $\dictnk{n}$, we consider lengths $n \in \{100, 200, 300, 500, 700\}$. We experiment with various rates of noise (10 - 30\%). While LSTMs do generalize well up to certain lengths, they achieve poor validation accuracy (<75\%) as the lengths go higher. At the same time, they obtain $100\%$ training accuracy on all the datasets. The validation accuracies of LSTMs are reported in Figure \ref{fig:rnn_majdict}. In contrast, Transformers achieve near-perfect generalization even in the presence of significant noise.

\textbf{Random k-sparse functions.} For $\randkn$, we experiment with various datasets for $\rand{n}{5}$ with $n \in \{30, 50, 80, 150, 200\}$. For lengths $n < 150$, we find that LSTMs generalize well on some of the $\rand{n}{5}$ functions. However, in the presence of $10\%$ noise (i.e., $\eta=0.1$), their performance degrades sharply. We create 10 datasets for $\rand{50}{5}$ with $\eta=0.1$, and similar to previous scenarios, LSTMs struggle to generalize well (>75\%) whereas Transformers are able to generalize perfectly on all the datasets (see Figure \ref{fig:intro}, top middle). However, even when the validation accuracies of LSTMs were below $75\%$, their training accuracy reached $100\%$ indicating that they overfit on the training data. Figure \ref{fig:intro} (bottom left) shows the training and validation curves of LSTMs on the $10$ datasets.

\textbf{Sensitivity During Training.} We observe that on $\kspar$ functions, both Transformers and LSTMs learn functions of increasing sensitivity. However, when LSTMs overfit and reach zero training error, they converge to functions of much higher sensitivity than that of the target function (see Figure \ref{fig:sensi_ksparse}. Since Transformers generalize perfectly, their sensitivity matches that of the target function.

\section{Clarifications}\label{app:clarifications}

\textbf{(1)} \textit{Do our results imply that Transformers can learn any $\kspar$ functions with small (practical) number of examples?} No. For small lengths ($n=50$) and $k=3$, we could enumerate and verify that they are able to learn all functions in the presence of 10\% noise. However, as the length $n$ and the number of relevant bits $k$ grow, Transformers struggle to perform well. Given the computational hardness associated with learning $\spar$, the task becomes much more difficult with the increase in $n$ and $k$. For $n=100$ and $k=5$, we were not able to obtain good generalization performance with Transformers.

\smallskip

\noindent \textbf{(2)} \textit{Do Transformers never overfit on $\kspar$ functions?} They do overfit when the size of the training data is very small. For $\spar$ with $n=40$ and $k=4$, it is perhaps surprising that Transformers learn the correct function even with as little as $2500$ training examples in less than $10000$ computational steps. However, for training sets of size $1000$, Transformers overfit across all runs. Additionally, with training sets of size $5000$ - $10000$, Transformers with higher depths overfit in some cases. See Appendix \ref{app:parities} for more details.

\smallskip

\noindent \textbf{(3)} \textit{Does the low sensitivity bias of Transformer (Section \ref{sec:sensi_exp}) explain their good generalization performance on $\kspar$ functions such as $\spar$?} No. Our findings in Section \ref{sec:sensi_exp} motivated us to compare the performance of Transformers and LSTMs on functions of low sensitivity such as $\kspar$ functions. While the bias towards low sensitivity functions and strong performance on various $\kspar$ functions could be related, it is not a direct explanation for their performance on $\kspar$. For $\spar$, it is natural to expect Transformers to follow some mechanism along the lines presented in \citet{hiddenprogressdl} for FFNs trained with SGD. However, the exact details are unclear, and more importantly, why and how LSTMs overfit is unclear as well.

\smallskip

\noindent \textbf{(4)} \textit{Are Transformers performing better than LSTMs because of learnable positional embeddings?} This seems unlikely since we found that Transformers with absolute positional encoding also generalize well on sparse parities (see Figure \ref{fig:san_grok}). Moreover, we found that LSTMs with learnable positional embeddings also fail to generalize on sparse parities and behave similarly to Figure \ref{fig:sparity_results}.

\smallskip

\noindent \textbf{(5)} \textit{Do LSTMs never succeed in learning $\spar$ from data?} They do succeed for smaller lengths. For lengths up to 20, we find that both Transformers and LSTMs are able to learn $\pari$ and $\spar$. However, for higher lengths, Transformers struggle to fit $\pari$ and LSTMs begin to overfit on $\spar$. For length $n=20$ and $k=4$, we could robustly find that even LSTMs without positional embeddings succeeded in learning sparse parities. On the other hand, for $n=40$ and $k=4$, we robustly found that LSTMs with learnable positional embeddings overfit and achieve poor generalization performance. Transformers were able to generalize well in the presence of noise across various hyperparameters for $\spar$ with $n=40$ and $k=4$. Our goal is not to identify the exact class of functions that Transformers can learn in practice. The key result is the juxtaposition of the performance between Transformer and LSTMs across various $\kspar$ functions.

\noindent \textbf{(6)} \textit{Do Transformers work effectively in practice primarily due to their simplicity bias?} It is hard to answer this question. In our work, we try to highlight concrete differences between Transformers and LSTMs with respect to certain properties which have close connections to generalization. While these properties could partially be the reason behind their good generalization performance, it is also possible that they are ubiquitous in practice because they effectively model long-distance dependencies and can be trained efficiently.

\section{Discussion and Final Remarks}\label{sec:discussion}
 A natural question that arises from our results is whether Transformers are performing better because the tasks are more suited to their architecture. Perhaps yes. One could argue that a number of regular languages that Transformers struggle to learn \citep{bhattamishra-etal-2020-ability, deepminddeletang2022neural} are more suited to recurrent architecture. Transformers have been shown to perform poorly on languages that require modular counting. DFAs, which are often considered to be formal abstractions of recurrent models, can represent these more efficiently. For instance, languages like standard parity can be represented with a two-state DFA while representing sparse parities would require a larger number of states. In contrast, for circuits that have recently been related to Transformers \citep{hao-etal-2022-formal-circuits, merrill-etal-2022-saturated-circuits}, representing sparse parities would be easier than representing standard parity. Our results indicate that previous works might have overestimated the performance of LSTMs by considering regular languages which are more suited for autoregressive architectures. 

 The question of which formal languages are more closely associated with practical tasks is not entirely clear. Prior works on analysis with formal languages have primarily followed Chomsky hierarchy owing to the conjecture that natural languages are mildly context-sensitive. While regular languages such as $\pari$ have high sensitivity ($\sensi=1$), practical tasks are often structured and have typically much lower sensitivity \citep{hahn-etal-2021-sensitivity}. In tasks such as sentiment analysis, the label often depends on a sparse subset of input tokens. When practical text datasets such as SST are labelled with random noise, then it can be shown that their sensitivity would be concentrated around $1/2$. As shown in Fig. \ref{fig:sst_noise}, models take much longer to fit such datasets whereas, in the case of the true labels, they only need a few epochs to fit the dataset.

 Our results indicate that while Transformers perform poorly on certain regular languages, they generalize more effectively than recurrent models on various sparse Boolean functions. Moreover, we showed that random Transformers as well as those trained with gradient-based algorithms are biased towards functions of low sensitivity. Our results add to the body of work that suggests that there is a form of implicit regularization in the procedure used to train neural models which prevent them from overfitting despite their incredible capacity.

\section*{Acknowledgments}

We would like to thank Michael Hahn, Ard Louis, Kabir Ahuja, anonymous reviewers, and our colleagues at the University of Oxford for helpful discussions and for providing valuable feedback.

\section*{Limitations}
A general limitation of this line of work is that most of the results are primarily confined to artificial datasets. Although such formal languages provide us with a controlled setting and clarity regarding the precise nature of the problem, the relation to practical tasks remains unclear. Hence, while our results highlight the contrast in the performance between the two types of architectures, its precise implications on real-world tasks remain unclear.

There are two negative results that do not support our hypothesis. (a) All the experiments discussed in the main paper are on strings of fixed lengths. We conducted some experiments on tasks with variable length sequences which in some sense have low sensitivity. The tasks can be seen as a variable length extension of sparse parities and sparse majorities. Unlike the fixed length setting, we found both LSTMs and Transformers perform similarly on those tasks. See Section \ref{app:varlen} in the Appendix for more details. (b) Although we found Transformers to consistently converge to low sensitivity functions in the case of Boolean functions, we did not find similar behaviour on sentiment classification datasets such as SST and IMDB (see Section \ref{app:sensi_real}).

A caveat with empirical studies such as this is that the results depend on the hyperparameters and other aspects of the experimental setup. While we have tried to be as thorough as possible with hyperparameter tuning, there is always a chance that the results or behaviour could differ for some hyperparameter.

\section*{Ethics Statement}
We have extensively discussed the limitations of our work in the previous section. We use two existing datasets, SST \cite{sst} and IMDB \cite{imdb}, which are publicly available and commonly used in NLP research. We synthetically generate datasets of formal languages which does not require ethical consideration. We have discussed the experimental details and computational budget in detail in Appendix \ref{app:implementation}. The research presented in this paper focuses on analysing the inductive biases of Transformers and LSTMs based on experiments on formal languages and subsequently we believe that our work does not raise any ethical concerns.

\bibliography{citations}
\bibliographystyle{acl_natbib}

\clearpage
\newpage

\appendix

\section{Roadmap}\label{app:roadmap}

The appendix is organized as follows. 

\begin{itemize}

    \item In Section \ref{app:randcomp}, we report and discuss additional results on the complexity of random models.
    
    \item In Section \ref{app:sensi_real}, we investigate the sensitivity of models on real data. In particular, we demonstrate that models learn functions of increasing sensitivity on sentiment classification datasets such as SST and IMDB.
    
    \item In Section \ref{app:sensi_gen}, we discuss some additional results relating sensitivity and generalization.
    
    \item In Section \ref{app:parities}, we present additional experiments investigating the ability of Transformers and LSTMs to learn sparse boolean functions.
    
    \item In Section \ref{app:rand_label}, we present some experiments to show that both Transformers and LSTMs can easily fit practical datasets even when they are labelled randomly.
    
    \item In Section \ref{app:implementation}, details of implementation and experimental setup are discussed which are relevant for the reproducibility of the results.
    
    \item In Section \ref{app:relwork}, we discuss some additional works related to our paper. 
    
\end{itemize}

\section{Complexity of Random Models}\label{app:randcomp}

In this section, we discuss additional results related to the complexity of random Transformers and LSTMs. We present results with additional complexity measures, initialization strategies, and variations across hyperparameters.

\subsection{Additional Measures}\label{appsub:complexity}

As discussed in Section \ref{sec:background}, sensitivity is related to several other complexity measures. Since it is more tractable to estimate sensitivity as opposed to certain other measures, we primarily focused on estimating and comparing sensitivity in the main paper. We explore three other complexity measures which have been previously explored in the literature to compute the complexity of functions represented by neural models. The measures are defined as follows:

\begin{enumerate}
    \item \textit{\textbf{SOP} (Size of Boolean Expression)}: This measure computes the size of the smallest Boolean expression in Sum-of-Product form that represents the function. In order to compute this for a neural network over $\hamn$, we compute the output of the model over all $2^n$ inputs and then use standard libraries (SymPy \citep{sympy}) to find the Boolean expression. The size indicates the number of operators and operands in the smallest expression. Since the problem of minimizing Boolean expressions is NP-complete, the runtime grows exponentially, and hence, we can only compute this up to length $10$ for several samples of random models. This measure was explored in \citet{valle-perez2018deep}.
    
    \item \textit{\textbf{Entropy}}: This measure takes the output labels for all $2^n$ inputs and simply computes the entropy over the labels. This is a weak measure and primarily indicates how imbalanced the label set is. This measure was explored in \citet{mingard2019neural, valle-perez2018deep}. 
    
    \item \textit{\textbf{CSR} (Critical Sample Ratio) }: This measure computes the fraction of inputs for which the function label changes at a small fixed distance from the inputs \citep{arpit_memorization_icml}. For discrete inputs such as $\hamn$, CSR can be seen as the fraction of inputs for which the function label changes at a Hamming distance of $1$. This was also explored in \citet{valle-perez2018deep}.

\end{enumerate}

\begin{figure}[t]
\centering
   \includegraphics[scale=0.4]{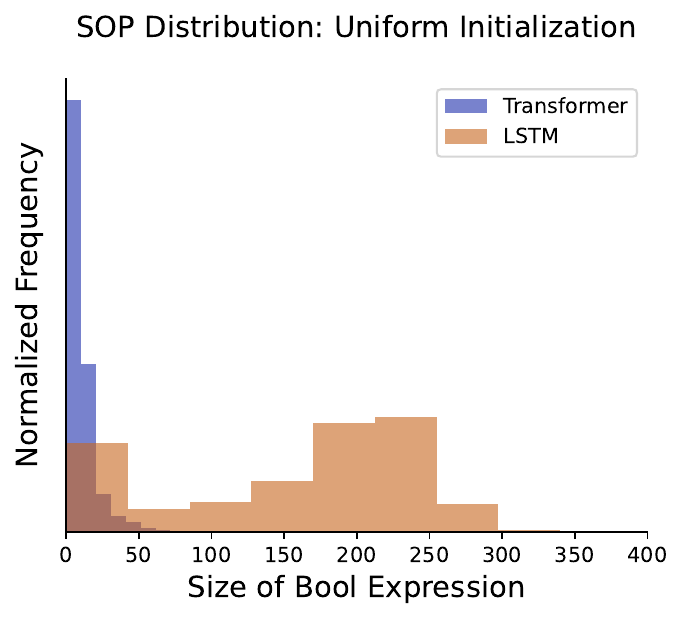}
	\caption{Distribution of SOP measure based on 50k samples of 4-layer Transformers and 1-layer LSTMs with width 64. \label{fig:sop_overlap} Refer to Section \ref{appsub:complexity} for details. }
	
\end{figure}

\begin{figure*}
\centering
\begin{subfigure}[b]{0.65\textwidth}
   \includegraphics[width=1\linewidth]{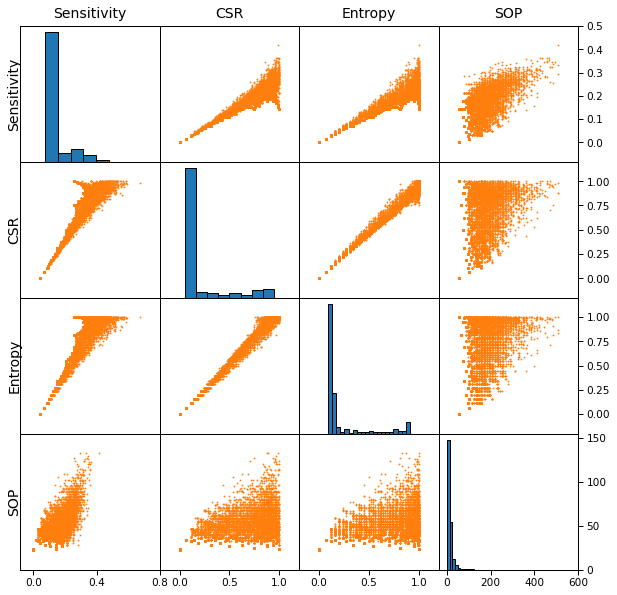}
   \caption{Transformers}
   \label{fig:comp_san} 
\end{subfigure}

\begin{subfigure}[b]{0.65\textwidth}
   \includegraphics[width=1\linewidth]{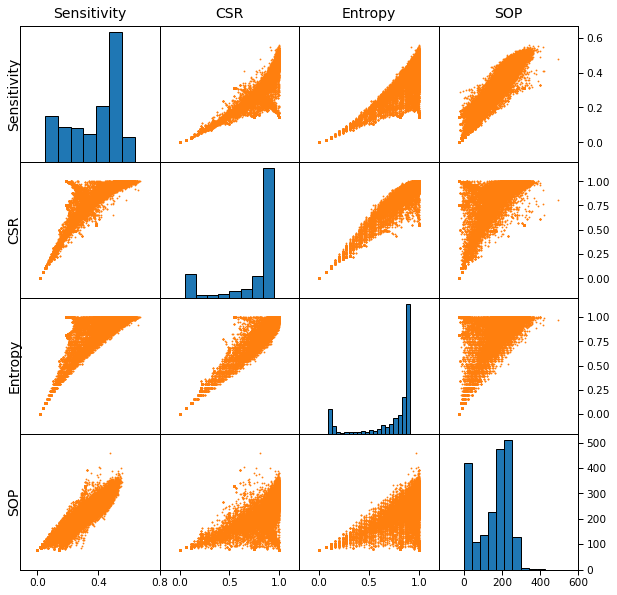}
   \caption{LSTMs}
   \label{fig:comp_lstm}
\end{subfigure}

	\caption{\label{fig:rand_comp} The distributions (diagonals) and comparisons (off-diagonals) of 4 different complexity measures for random (a) Transformers and (b) LSTMs with weights uniformly sampled in [-10, 10]. The measures are computed over all inputs in $\{0,1\}^7$. CSR denotes critical sample ratio and SOP denotes the size of the smallest expression in sum-of-product form that represents the truth table. Refer to Section \ref{appsub:complexity} for details. }
	
\end{figure*}

Figure \ref{fig:rand_comp} shows the distribution of different complexity measures and scatter plots depicting relations among them. The measures are computed for random Transformers and LSTMs with weights sampled uniformly between -10 and 10. The measures are computed for sequences of length $7$ with $200k$ samples of models. We take Transformers and LSTMs with depth $\in \{1, 2, 4, 8\}$ and width (d\_model/hidden\_size) $\in \{8, 32, 64, 256, 768\}$. We take an equal number of samples for each hyperparameter. Figure \ref{fig:sop_overlap} shows the distribution of SOP based on 50k samples for a fixed hyperparameter configuration of Transformer and LSTM. It includes a 1-layer LSTM with width $64$ and a 4-layer Transformer with width $64$.

As can be seen in Figure \ref{fig:rand_comp}, there exists significant correlation between sensitivity and other measures. Note that, high sensitivity functions will always have high entropy and high CSR but the converse is not true. Functions with maximum entropy can also have low sensitivity. For instance, the dictator function has maximum entropy (since half the inputs have label 1 and the other half have label 0) while having very low sensitivity. Similarly, CSR can be seen as a weaker version of sensitivity.

\subsection{Why Sensitivity?}\label{subsec:why_sensi}

Sensitivity can be seen as a discrete analog \citep{gopalan2016smooth} of the `smoothness' of a continuous function which measures how gradually a function changes locally. Functions of higher sensitivity can be considered more complex since the function value can be changed by changing any of a large subset of bits whereas functions of lower sensitivity depend on fewer bits and their function value can be determined based on a small number of input coordinates. Sensitivity measures are also polynomially related to several other notions of complexity such as the depth of a decision tree, certificate complexity, and the degree of the Fourier expansion of Boolean functions (see \citet{sensi_other_2014tighter} for more details). The correlation between generalization and a different notion of sensitivity has been demonstrated in \citet{novak2018sensitivity} for computer vision models. The relation between generalization and a variant of Boolean sensitivity has even been explored over a decade ago by \citet{franco2006generalization}.  More recently, \citet{hahn-etal-2021-sensitivity} extend the notion of block sensitivity to incorporate variable length sequences and propose it as a measure to estimate the difficulty of various NLP tasks.

\subsection{Additional Sensitivity Results}\label{appsub:addsensi}

\begin{figure}[t]%
	\centering
	\subfloat{{\includegraphics[scale = 0.29, trim=0 0 5 5, clip]{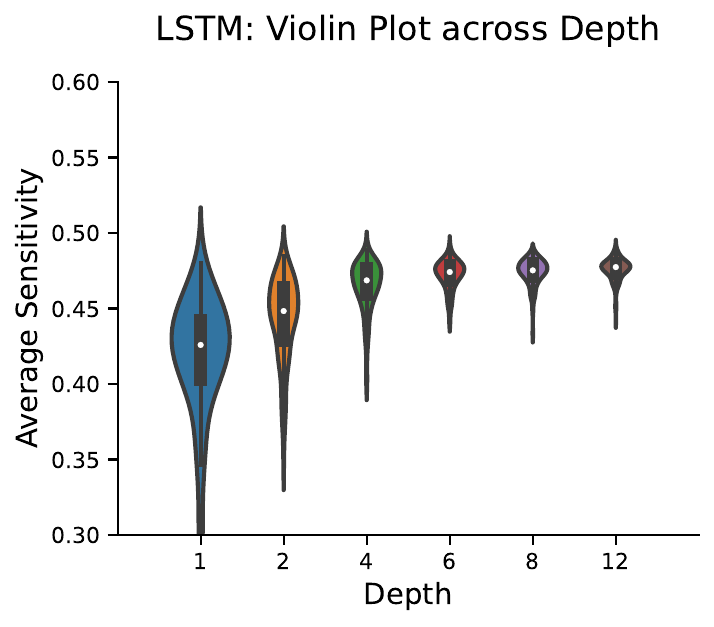} }}%
	\subfloat{{\includegraphics[scale = 0.29, trim=0 0 5 5, clip]{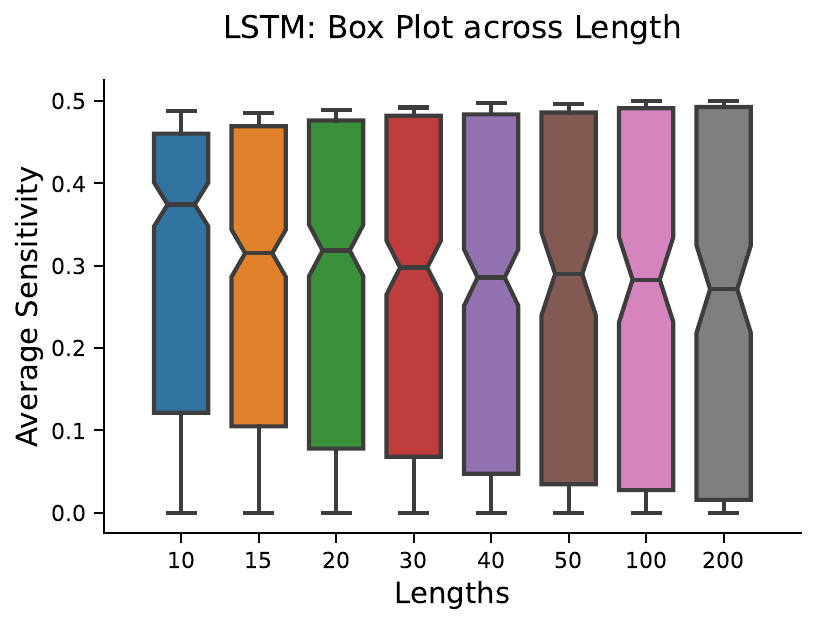}  }}
	
	\caption{ \label{fig:sensi_lstm_hyperparams} Mean sensitivity of Random LSTMs across different widths and lengths.  Refer to Section \ref{appsub:addsensi} for details. }%
	
\end{figure}

\begin{figure*}[t]%
	\centering
	\subfloat{{\includegraphics[scale = 0.35, trim=0 0 5 5, clip]{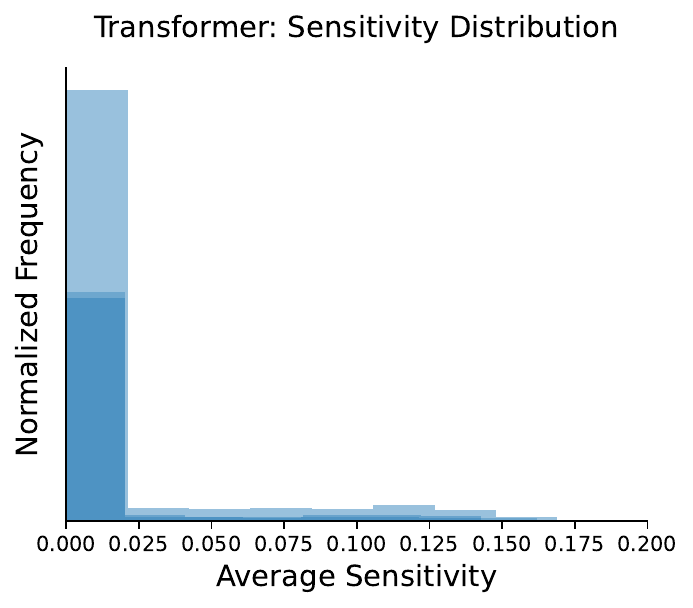} }}%
	\subfloat{{\includegraphics[scale = 0.35, trim=0 0 5 5, clip]{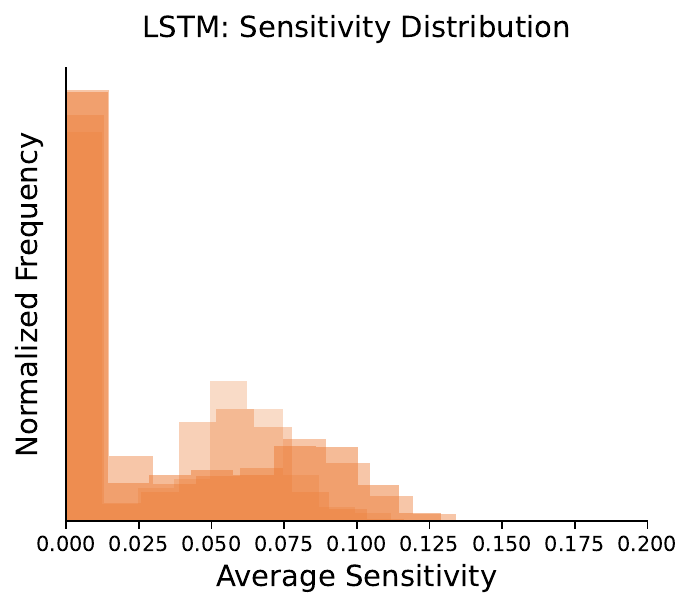} }}
	\subfloat{{\includegraphics[scale = 0.35, trim=0 0 5 5, clip]{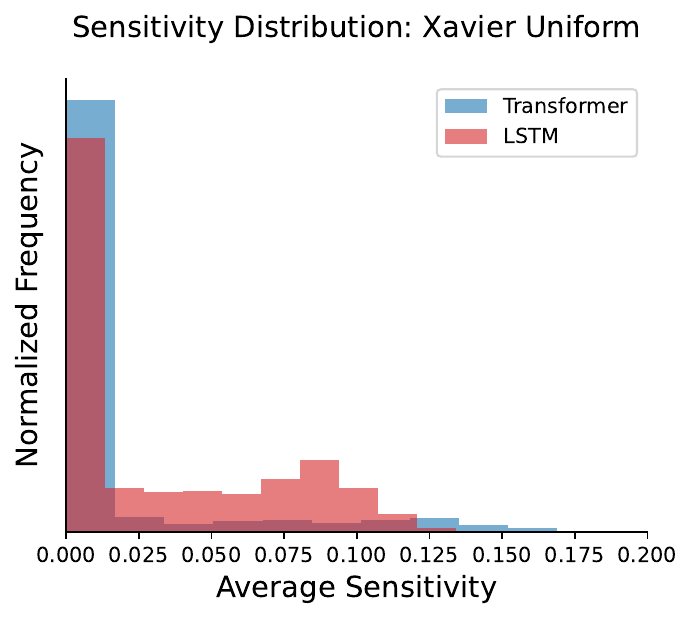} }}

	\caption{ Distribution of sensitivity of different Transformers and LSTMs with Xavier uniform initialization. Sensitivity of Transformers (left) and LSTMs (middle) across various hyperparameters. Right: Comparison for a fixed hyperparameter. Refer to Section \ref{appsub:addsensi} for details.}%
	\label{fig:xav_uni}%
\end{figure*}

The distribution of the sensitivity of Transformers and LSTMs initialized with Xavier uniform distribution are given in Figure \ref{fig:xav_uni} respectively. For Gaussian initialization, the weights are sampled with mean 0 and $\sigma=10$. For Xavier uniform initialization all the values in weight matrices are sampled uniformly between $-d^{1/2}$ and $d^{1/2}$ where $d$ is the number of hidden units. The values in the bias vectors are set to zero and the ones in the input embedding vectors are sampled from $\mathcal{N}(0,1)$.

\textbf{Finding Parity.} For strings of length $5$, the total number of possible functions is $2^{2^5}$. If Boolean functions are sampled uniformly, then the probability of picking the $\pari$ function is less than 1 in two billion. However, on uniformly sampling $10$ million LSTMs of depth $2$ and hidden size $8$, we found that the probability of finding one that represents $\pari$ is 1 in 30,000. Hence, it is over \textbf{60,000} times more likely to find $\pari$ function by sampling LSTMs than randomly sampling Boolean functions. This indicates that the parameter space of recurrent models such as LSTMs has a significant representation of $\pari$ functions which might help explain why it is easier for them to learn $\pari$. On the other hand, for Transformers, we did not find a single sample which represented $\pari$ based on 10 million samples.

\begin{figure*}[t]%
	\centering
	\subfloat{{\includegraphics[scale = 0.35, trim=0 0 5 5, clip]{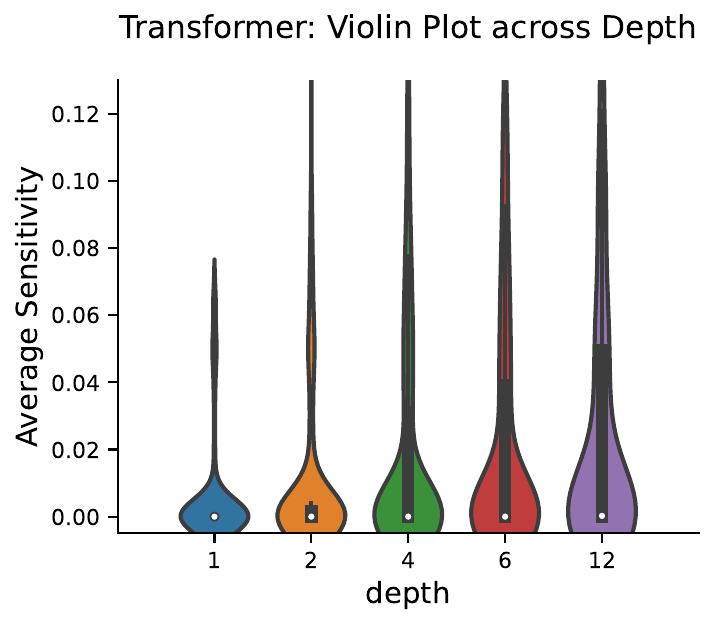}  }}%
	\subfloat{{\includegraphics[scale = 0.35, trim=0 0 5 5, clip]{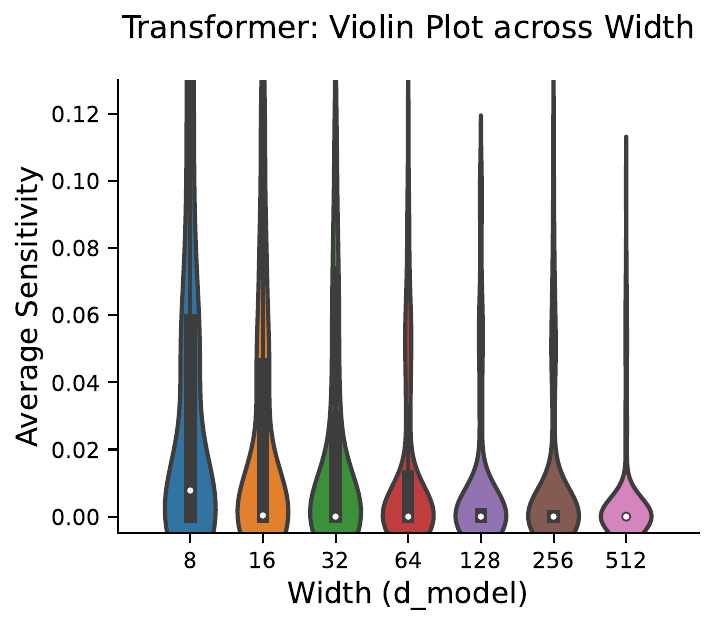}   }}
	\subfloat{{\includegraphics[scale = 0.35, trim=0 0 5 5, clip]{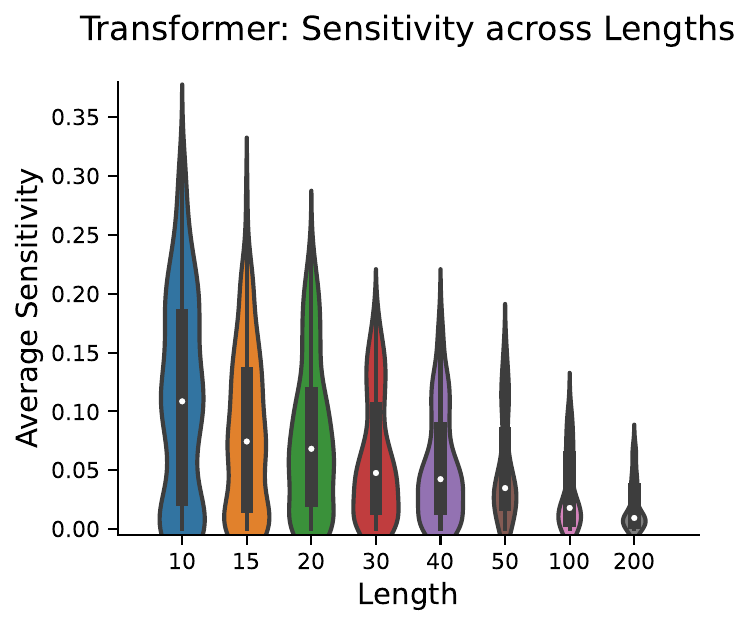} }}

	\caption{ \label{fig:sensi_hyperparams}  Sensitivity of random Transformers across different depths, widths, and lengths.  Refer to Section \ref{subsec:randsensi_exp} for details.}%
	
\end{figure*}

\textbf{Change across hyperparameters.} For uniform sampling, a general observation for both the architectures is that the likelihood of higher sensitivity functions increases with the number of layers (see Figure \ref{fig:sensi_hyperparams}, left and Figure \ref{fig:sensi_lstm_hyperparams}), however, even for Transformers with depth 12, the distribution is heavily skewed towards low sensitivity functions in comparison to recurrent models with depth 1. Unlike recurrent models, the sensitivity of Transformers decreases when the width of the model is increased (see Figure \ref{fig:sensi_hyperparams}, middle). For Transformers, the average sensitivity decreases with the increase in the length of the strings (see Figure \ref{fig:sensi_hyperparams}, right), whereas for LSTMs, it remains quite high even for lengths up to 200.

For LSTMs with uniform sampling, the change in sensitivity across different widths (hidden\_size) and lengths is provided in Figure \ref{fig:sensi_lstm_hyperparams}. As can be seen, the sensitivity of LSTMs does not significantly reduce across higher lengths and widths, unlike Transformers.

While it is not entirely clear why random Transformers are relatively more biased towards low complexity functions, we observe that they behave similar to hard-attention Transformers upon inspection of attention weights. Recent works \citep{hao-etal-2022-formal-circuits, hahn-2020-theoretical} have shown that hard-attention Transformers can only represent functions in $\text{AC}^0$ (which contain functions that can be represented by constant depth AND/OR circuits). Since $\text{AC}^0$ circuits can only represent functions of low average sensitivity \citep{bool_analysis}, it might help explain why random Transformers have low sensitivity.

\section{Sensitivity During Learning Sentiment Classification}\label{app:sensi_real}

In this section, we discuss experiments on measuring the sensitivity of Transformers and LSTMs when trained on the sentiment classification task.

\subsection{Experimental Setup}

\textbf{Datasets.} We experiment with two sentiment classification datasets: SST \citep{sst} and IMDB \citep{imdb}. For SST, we train on the full train set of size 67349 examples and evaluate both sensitivity and validation accuracy on the validation set of size 872 examples. For IMDB, we preprocess the dataset to only include sentences of length up to 500. This leads to a train set of size 22156. The validation set consists of 8939 examples randomly sampled from the test set. Since the sentences in IMDB dataset are of much longer lengths, in order to save compute, we evaluate sensitivity of models on a dataset of size 894 examples randomly sampled from the test set.

\textbf{Sensitivity Metrics.} Boolean sensitivity as defined in Section \ref{sec:background} cannot be directly applied to sequences of variable length and larger vocabulary. As an alternative, we compute certain proxy metrics which measure how likely it is for the function value to change due to a change in one token of the input sequence. To that end, we design three simple metrics to measure the sensitivity of models trained on sentiment classification:

\begin{enumerate}
    \item\emph{Word Label-Sensitivity:} For each word in the sentence (one word at a time), we replace it $n$ times with a word sampled randomly from the vocabulary and measure the average (over $n$) number of times the predicted label changes. We sum this value for all the words in the sentence and normalize the value by its length.

    \item\emph{Word Softmax-Sensitivity:} For each word in the sentence (one word at a time), we replace it $n$ times with a word sampled randomly from the vocabulary and measure the average (over $n$) $L2$-distance between the predicted softmax normalized output vector before and after the replacement. Again, we sum this value for all the words in the sentence and normalize by its length.

    \item\emph{Embedding Label-Sensitivity:} For each word in the sentence (one word at a time), we add Gaussian noise with mean $0$ and variance $\sigma^2$ to its embedding $n$ different times and measure the average (over $n$) number of times the predicted label changes. We sum this value for all the words in the sentence and normalize by its length.

\end{enumerate}

For all metrics, the final score is obtained by averaging across all the examples in the dataset. In all our experiments, we set $n = 10, \text{ and } \sigma^2 = 15$.

\begin{figure}[t]%
	\centering
	\subfloat{{\includegraphics[scale=0.25, trim=0 5 0 0, clip]{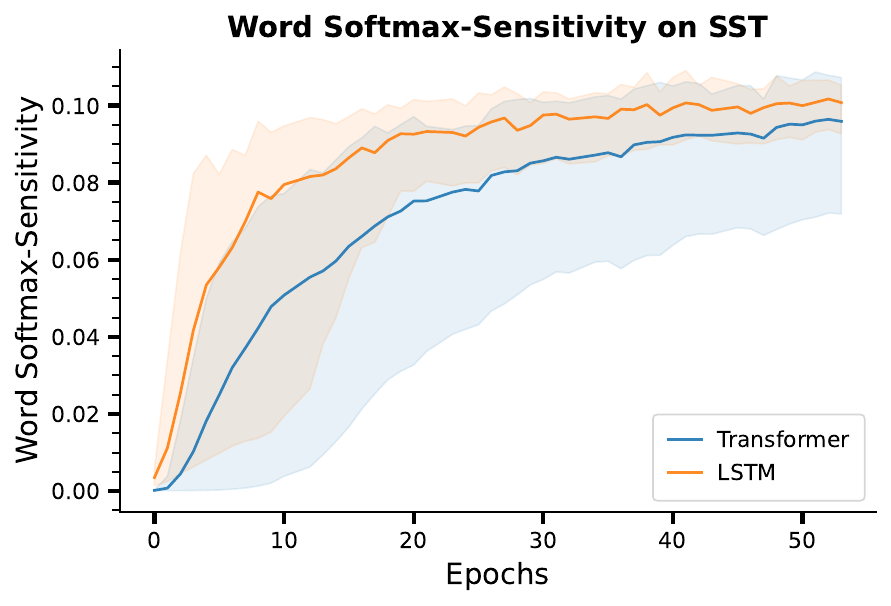} }}%
	\subfloat{{\includegraphics[scale=0.25, trim=0 5 0 0, clip]{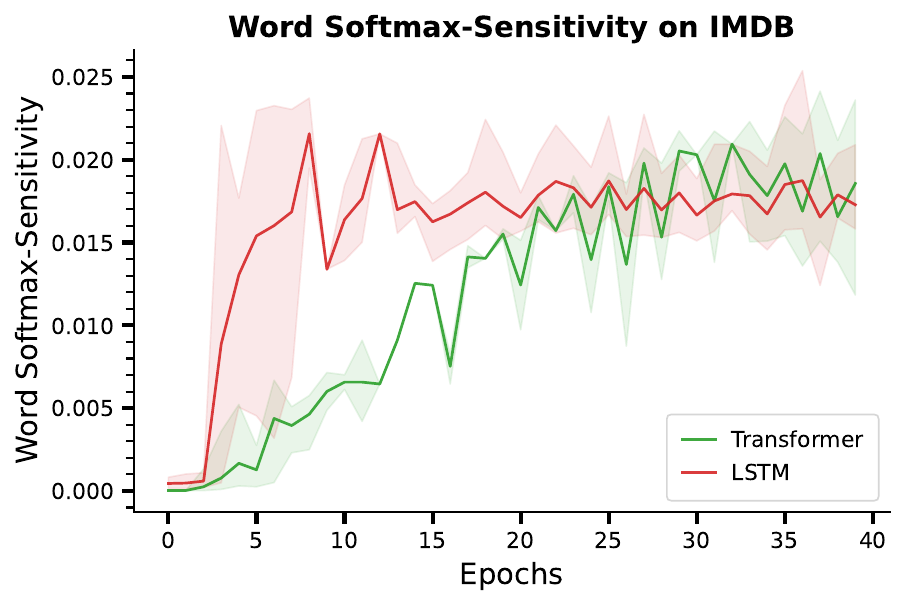}  }}

	\caption{\label{fig:sensi_imdb_sst} The `word softmax-sensitivity' of Transformers and LSTMs at different stages of training on SST and IMDB datasets.}%
	
\end{figure}

\textbf{Hyperparameter Details.} For both Transformers and LSTMs, we vary the number of layers $\in \{1, 2\}$, learning rate $\in \{0.0001, 0.0003, 0.0005\}$, and model width (d\_model/hidden\_size) $\{128, 256\}$. We set the batch size as $128$ and the FFN size as twice the width. For LSTMs, we keep the embedding size the same as the hidden size. Both models are trained with Adam \citep{adam} optimization and using Dropout regularization with probability $0.2$.

\begin{figure}[t]%
	\centering
	\subfloat{{\includegraphics[scale=0.25]{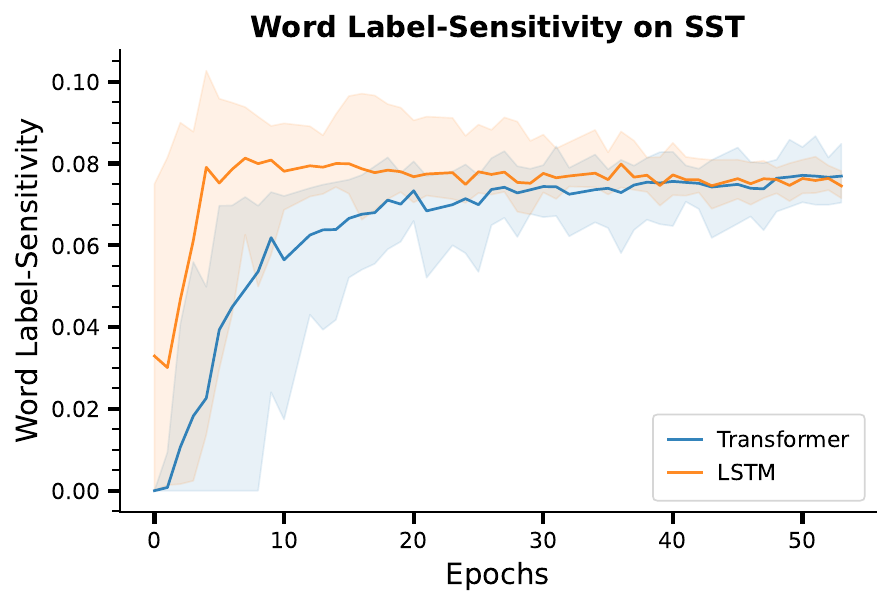} }}%
	\subfloat{{\includegraphics[scale=0.25]{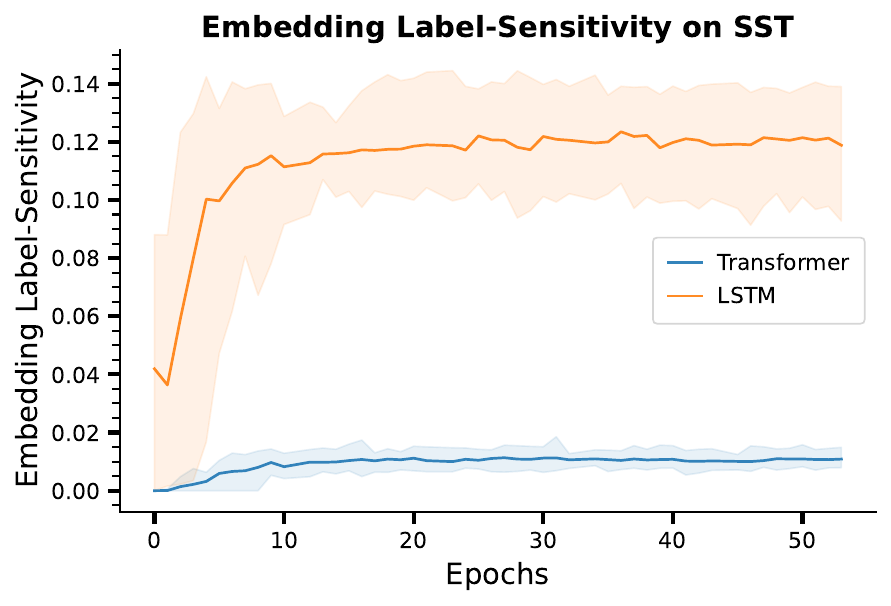}  }}


	
	\caption{\label{fig:sensi_sst} The `word label-sensitivity' and `embedding label-sensitivity' of Transformers and LSTMs at different stages of training on SST dataset.}%
	
\end{figure}

\textbf{Results.} Figure \ref{fig:sensi_imdb_sst} shows the word softmax-sensitivity for both models across different iterations of training for SST and IMDB datasets. The word label-sensitivity and embedding label-sensitivity for SST is provided in Figure \ref{fig:sensi_sst}.

\noindent We find that across all three measures, both Transformers and LSTMs learn functions of increasing sensitivity where they prioritize learning functions of lower sensitivity first. We found `word label-sensitivity' and `word softmax-sensitivity' to correlate well with \emph{generalization gap} (i.e., the difference between train accuracy and test accuracy). Since the measures are very similar, there is a strong correlation between the two measures themselves. We did not find any non-trivial correlation between `embedding label-sensitivity' and generalization gap. Note that unlike random and sparse Boolean functions, on real datasets, we did not find Transformers converging to functions with lower sensitivity. 

\section{Sensitivity and Generalization}\label{app:sensi_gen}

 \subsection{Sensitivity as Capacity Measure}

 We show how maximum sensitivity can be used as a capacity measure to derive generalization bounds.  Capacity measures such as the VC Dimension are a classical approach to derive sample complexities and probabilistic upper bounds for the test error of a classification model.  

Let $\funcset_k:\hamn \rightarrow \{\pm1\}$ be a class of functions such that the maximum sensitivity for any function $f \in \funcset_k$ is upper bounded by $k$ where $0\leq k \leq n$. Any function $f$ with a maximum sensitivity $k$ can be uniquely determined by its values on any Hamming ball of radius $2k$ in $\hamn$ \citep{gopalan2016smooth}. This can be used to upper bound the size of the function class $|\funcset_k| \leq 2^{\binom{n}{\leq 2k}}$. Since the VC Dimension (denoted $\vcd$) of a class of functions $\funcset$ is upper bounded by $\log|\funcset|$, we have that,

\begin{equation}\label{eq:vcd}
    \vcd(\funcset_k) \leq \binom{n}{\leq2k} \leq \binom{n+2k}{2k} 
\end{equation}
\[
\leq \left ( \frac{e(n+2k)}{2k} \right )^{2k} = \mathcal{O}(n^{2k})
\]

Let $f \in \funcset_k$ be a target function and $\hat{f} \in \funcset_k$ be a hypothesis produced by a learning algorithm. Let $L(\hat{f}, f) = \Ex_{x\sim\hamn}[\indicator[\hat{f}(x) \neq f(x)]]$ be the true error between $f$ and $\hat{f}$. Similarly, let $\hat{L}_S(\hat{f}, f) = \frac{1}{m}\sum_{i=1}^{m}\indicator[\hat{f}(x) \neq f(x)]$ be the empirical error on a sample set $S$. Then using Equation (\ref{eq:vcd}) and basic properties of VC dimension (\citet{mohri2018foundations}), we can upper bound the distance of the true error $L$ from the sample error $\hat{L}$ using maximum sensitivity. 

\begin{proposition}
For any $\delta >0$, with probability at least $1-\delta$, the following holds for any function $f, \hat{f} \in \funcset_k$, 

\begin{equation}
    L(\hatf, f) \leq \hat{L}_S(\hatf,f) + \sqrt{\frac{cn^{2k}\log\frac{2em}{cn^{2k}} + 8\log\frac{4}{\delta} }{m}}
\end{equation}
\end{proposition}

where $c>0$ is some constant. Functions with low maximum sensitivity can be learned with better sample efficiency. Functions with low average sensitivity can also be learned efficiently when the data generating distribution is uniformly distributed over the input (\citet{bool_analysis}, Sec 3.4).

\subsection{Sensitivity and Generalization Gap}

The correlation between sensitivity and generalization has previously been studied for networks trained on Boolean functions \citep{franco2006generalization} and image datasets \citep{novak2018sensitivity}. We examine the relation between simple variants of sensitivity described in Section \ref{app:sensi_real} and generalization.

\begin{figure}[t]%
	\centering
	\subfloat{{\includegraphics[scale=0.23]{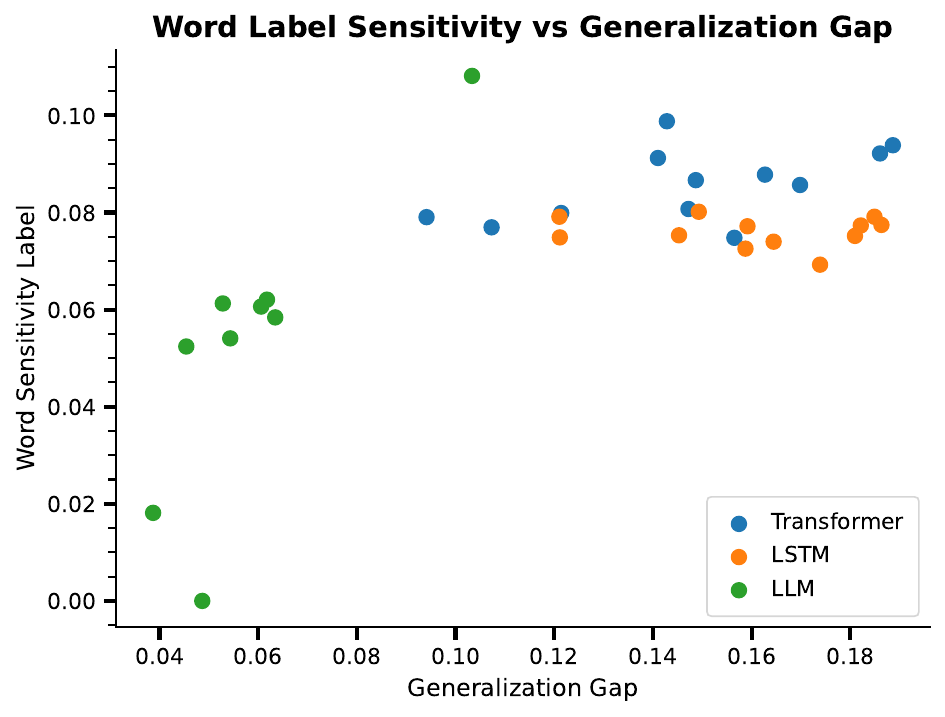} }}%
	\subfloat{{\includegraphics[scale=0.23]{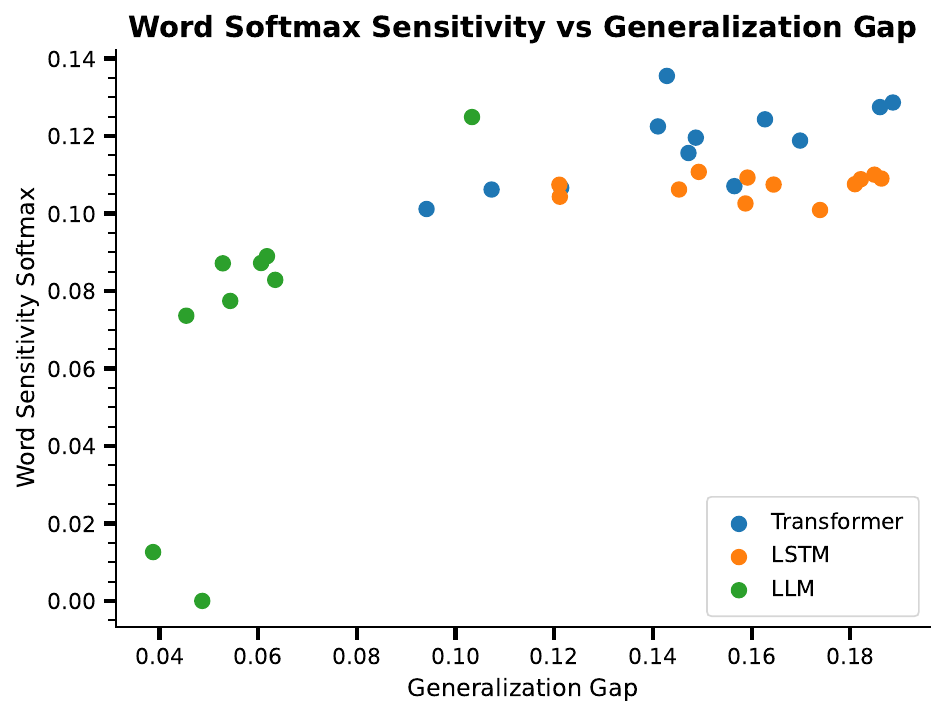}  }}
	
	\caption{\label{fig:sensi_gen_gap} Sensitivity vs Generalization gap for Transformers, LSTMs, and RoBERTa LLM trained on SST.}%
	
\end{figure}

We train various models on SST dataset until convergence and then compare sensitivity with generalization gap. The generalization gap is simply the difference between the train error and test error; higher gap indicates overfitting. We plot the word label-sensitivity and word softmax-sensitivity (defined in Section \ref{app:sensi_real}) for Transformers, LSTMs, and a pretrained Large Language Model (RoBERTa \citep{liu2019roberta}) against the generalization gap (see Figure \ref{fig:sensi_gen_gap}). We observe a positive correlation between the measures and generalization gap indicating that when sensitivity is higher, the models are more likely to overfit and achieve poorer generalization performance. Large language models such as RoBERTa have a lower sensitivity while achieving better test accuracies than Transformers and LSTMs trained from scratch.

\section{Additional Experiments on Sparse Boolean Functions}\label{app:parities}

\begin{figure}[t]%
	\centering
	\subfloat{{\includegraphics[scale = 0.25, trim=0 0 5 5, clip]{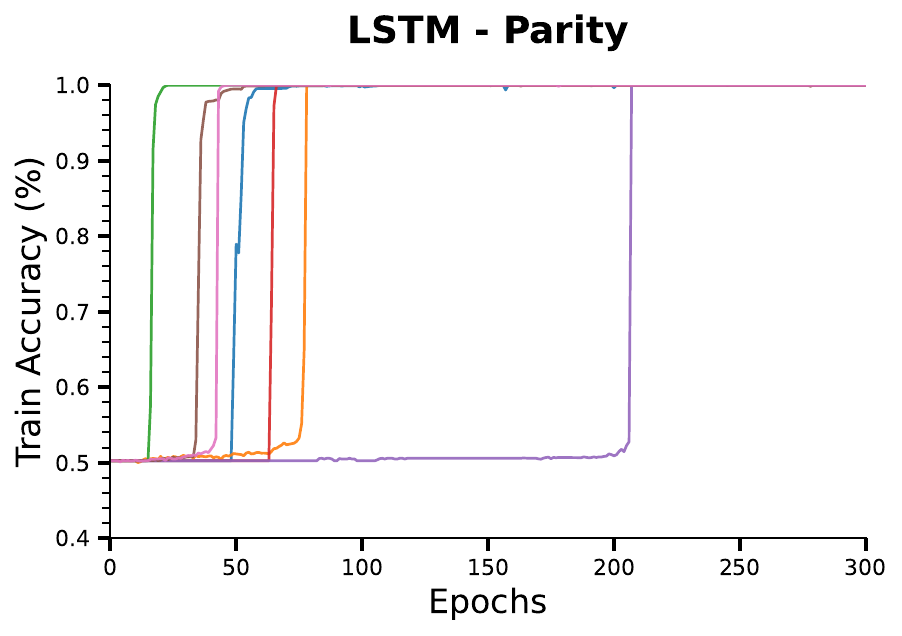} }}%
	\subfloat{{\includegraphics[scale = 0.25, trim=0 0 5 5, clip]{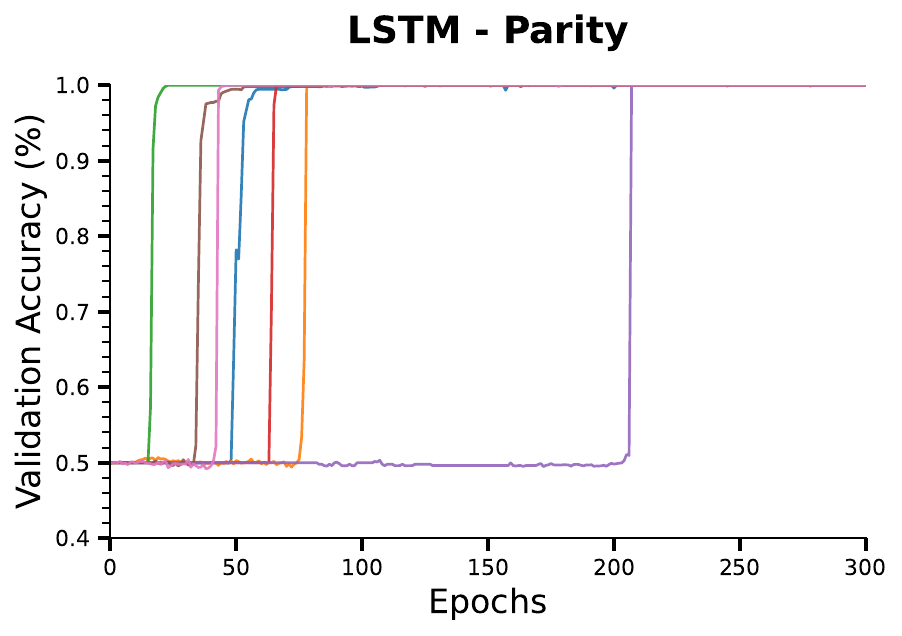}  }}
	
	\caption{\label{fig:rnn_parity} Training and validation curves for LSTMs trained on standard parity ($\parn{30}$) dataset containing 30k samples. Similar to the behaviour of Transformers on $\spar$, we observe abrupt phase transitions for LSTMs where there is a sharp increase in the train and validation accuracy.}%
	
\end{figure}

\textbf{Standard Parity.} The training curves for LSTMs on standard parity are provided in Figure \ref{fig:rnn_parity}. The models are trained on datasets of size 20k where the input strings are of length 30. Similar to Transformers on $\spar$, we observe phase transitions for LSTMs on standard $\pari$ task.

\medskip

\begin{figure}[t]%
	\centering
	\subfloat{{\includegraphics[scale = 0.25, trim=0 0 5 5, clip]{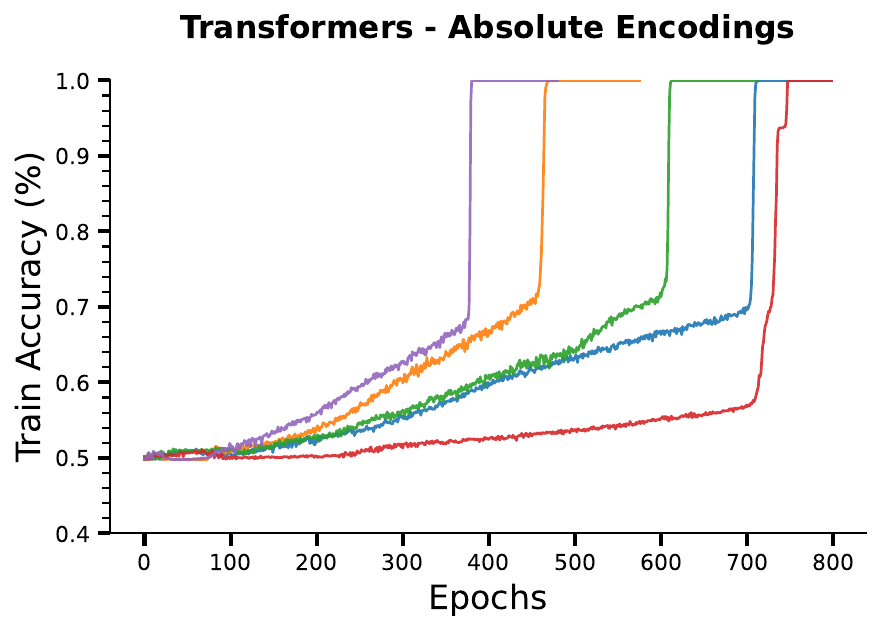} \label{fig:san_grok_tracc} }}%
	\subfloat{{\includegraphics[scale = 0.25, trim=0 0 5 5, clip]{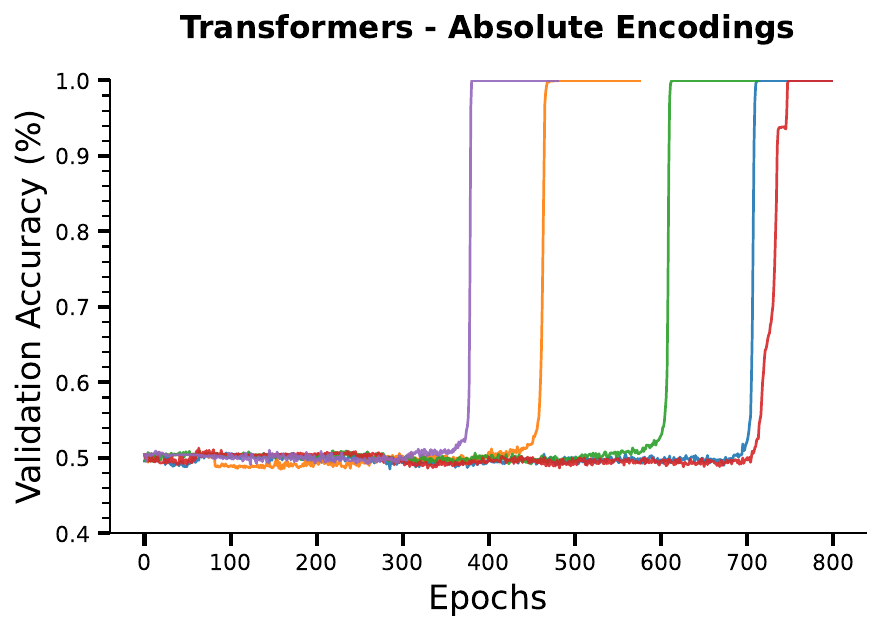} \label{fig:san_grok_valacc}  }}
	
	\caption{Training and validation curves for Transformers with absolute positional encodings trained on sparse parity task (length n=40 and k=4). The training accuracy gradually starts to increase with no change in validation accuracy. After a point, both training and validation accuracies match.}%
	\label{fig:san_grok}%
\end{figure}

\begin{figure}[t]%
	\centering
	\subfloat{{\includegraphics[scale = 0.24, trim=0 0 5 5, clip]{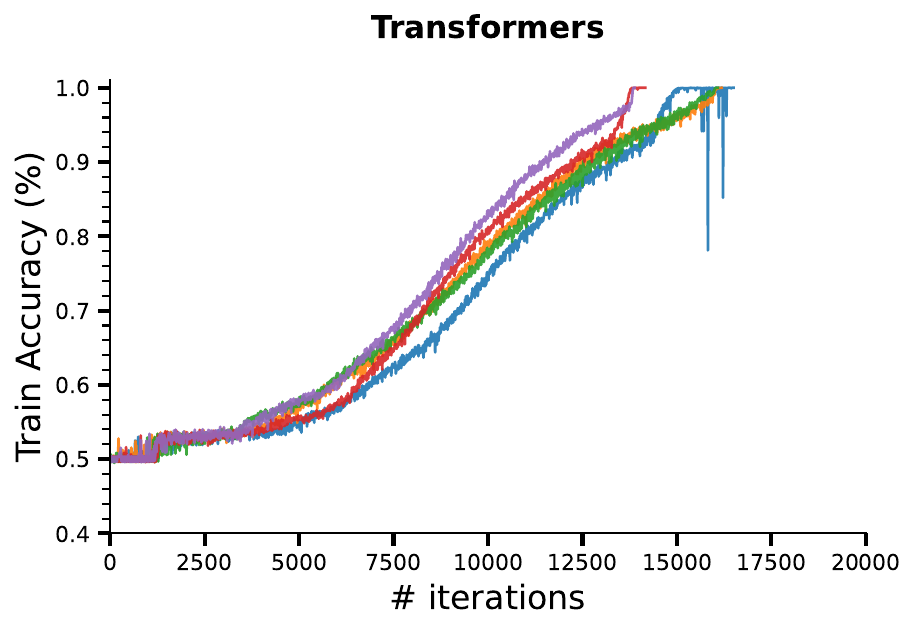} }}%
	\subfloat{{\includegraphics[scale = 0.24, trim=0 0 5 5, clip]{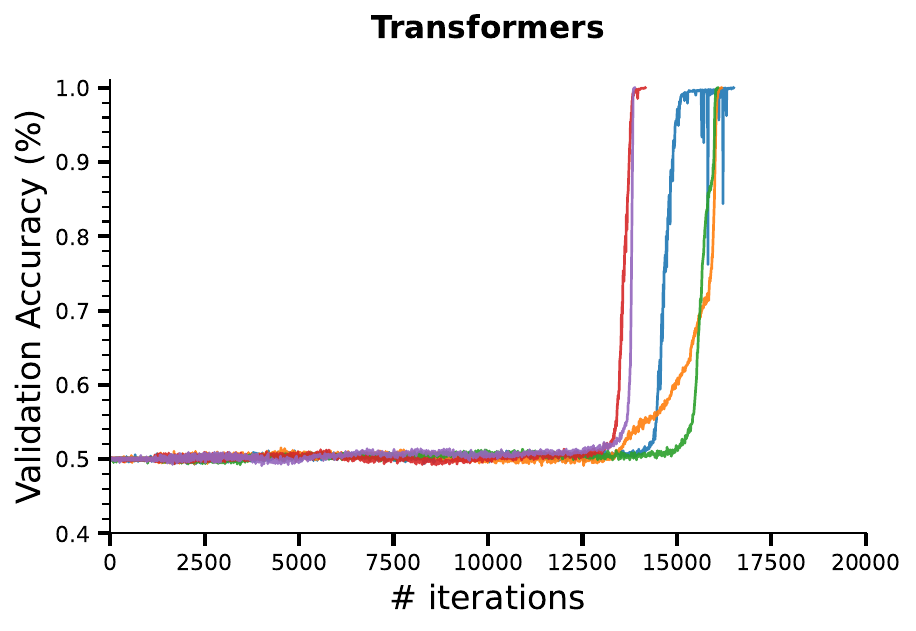}  }}
	
	\caption{Training and validation curves for Transformers trained on sparse parity dataset (length n=40 and k=4) with only 5k training examples. The training accuracy gradually starts to increase with no change in validation accuracy. After a point, both training and validation accuracies match.}%
	\label{fig:san5k_grok}%
\end{figure}

\noindent \textbf{Sparse Parities.} The results on sparse parities with length $n$=40 and $k$=4 relevant bits for Transformers with absolute positional encodings are provided in Figure \ref{fig:san_grok}. We find that Transformers with absolute positional encodings are able to generalize well on $\spar$ task and exhibit grokking on relatively larger datasets (30k samples) in comparison to models with learnable positional embeddings. For Transformers trained with learnable encoding, we robustly observe grokking on small datasets. Figure \ref{fig:san5k_grok} depicts the training curves for Transformers trained on datasets of size 5k.

\begin{figure}[t]
\centering
   \includegraphics[scale=0.4]{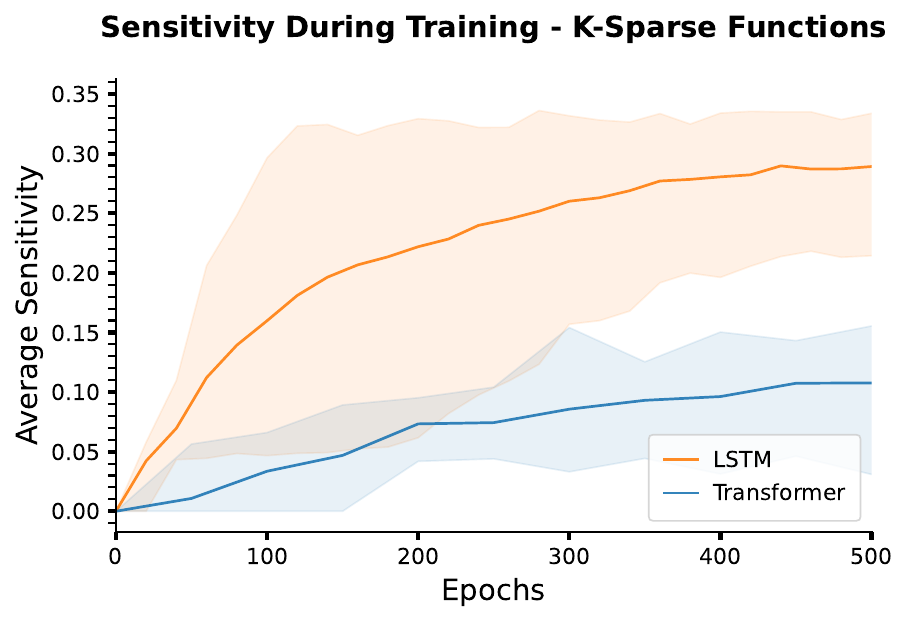}
	\caption{\label{fig:sensi_ksparse} The sensitivity of Transformers and LSTMs at different stages of training on various $\kspar$ functions of length $50$. Refer to Section \ref{sec:bool_exp} for details. }
	
\end{figure}

\medskip

\begin{figure}[t]%
	\centering
	\subfloat{{\includegraphics[scale = 0.25, trim=0 0 5 5, clip]{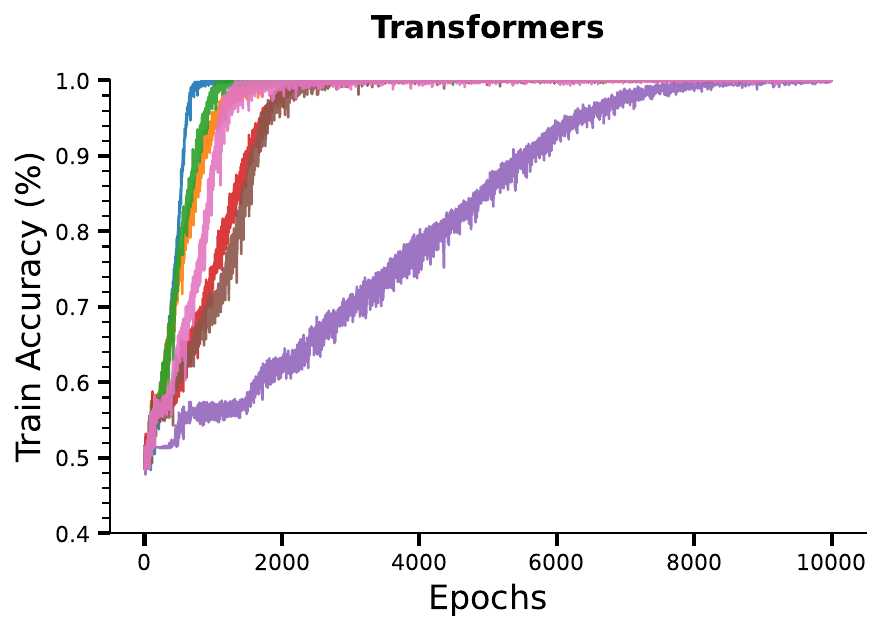} }}%
	\subfloat{{\includegraphics[scale = 0.25, trim=0 0 5 5, clip]{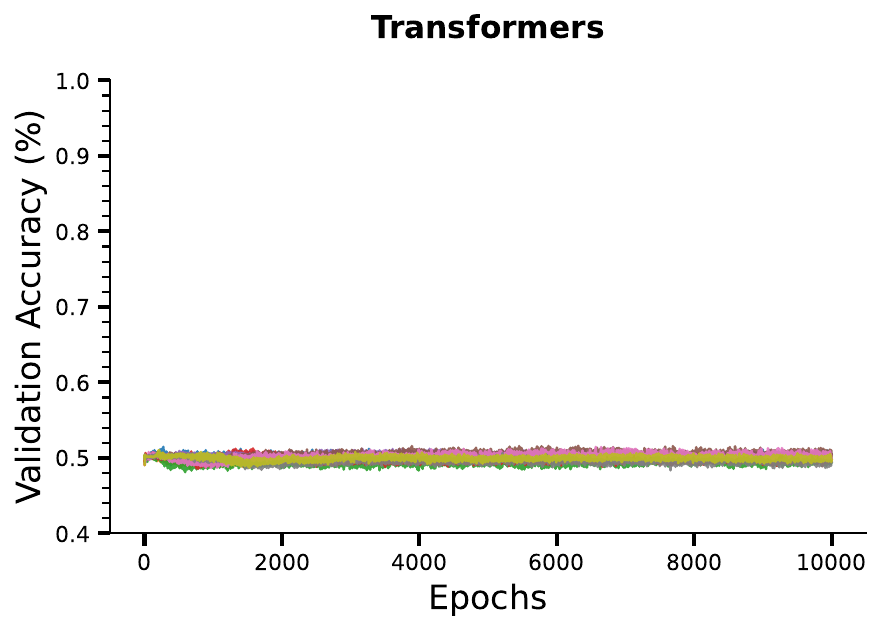}  }}
	
	\caption{Training and validation accuracy curves for Transformers trained on sparse parity dataset (length n=40 and k=4) with 1k training examples. The behaviour is similar to how LSTMs overfit for much larger training sets (40k samples). We observe similar behaviour for Transformer with large depths on datasets of size 5k.}%
	\label{fig:san_overfit}%
\end{figure}

\noindent \textbf{Overfitting.} We found that Transformers overfit on training data when the sample size is too low (see Figure \ref{fig:san_overfit}). Apart from that, for datasets of certain sizes, we find that while Transformers with depth up to $6$ generalize well, those with much higher depths ($> 8$) overfit across several runs.

\medskip

\noindent \textbf{Effect of Regularization on LSTMs.} We explore the effect of dropout and L2 regularization on training with LSTMs. While training on sparse parities, we find that increasing regularization increases the convergence time but the model still overfits and converges to a function with higher sensitivity than the target function. Upon further increasing regularization, the model fails to fit the training data. 

\medskip

\noindent \textbf{Mixed Parity.} To explore the difference in bias between Transformers and LSTMs, we conduct a simple experiment described as follows. We create a dataset of size 15k called `Mixed Parity' where half of the examples are labelled as standard $\pari$ (label is determined by all bits) and the other half is labelled as $\spar$ with $4$ relevant bits. The inputs are of length $30$ and the first bit determines whether the input is labelled according to standard $\pari$ function (when the first bit is 1) or as a $\spar$ function (when the first bit is 0). 

\begin{figure}[t]%
	\centering
	\subfloat{{\includegraphics[scale = 0.25, trim=0 0 5 5, clip]{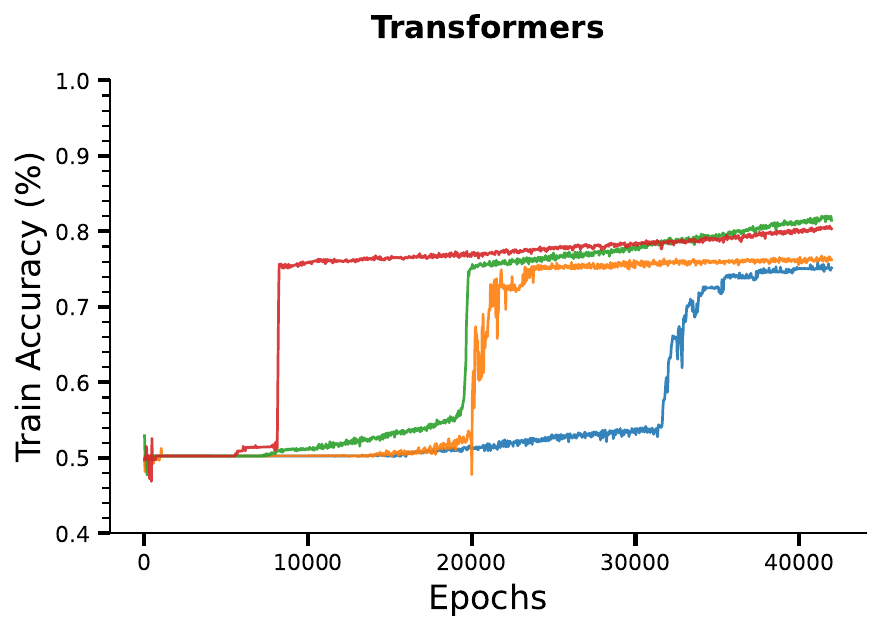} }}%
	\subfloat{{\includegraphics[scale = 0.25, trim=0 0 5 5, clip]{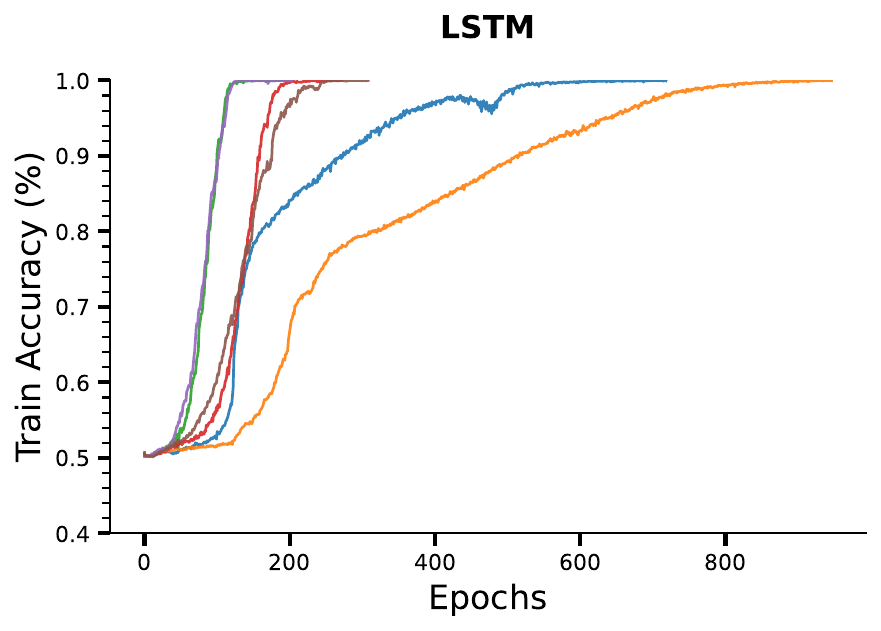}  }}
	
	\caption{Training curves for Transformers and LSTMs trained on the Mixed Parity dataset (length n=30) with 15k training examples. Refer to Section \ref{app:parities} for details.}%
	\label{fig:mixpar}%
\end{figure}

We train Transformers and LSTMs (with learnable positional encodings) of depth $2$ and width $64$ across various learning rates $\in [0.01, 0.00001]$ on the Mixed Parity dataset. We find that LSTMs obtain 100\% training accuracy on the dataset (see Figure \ref{fig:mixpar}, right); LSTMs validation accuracy on the $\pari$ task is near 100\% whereas it is 50\% on the $\spar$ task. In contrast, the training accuracy of Transformers converges around 75\% (see Figure \ref{fig:mixpar}, left); their validation accuracy on the $\pari$ task is 50\% whereas on $\spar$ they achieve near 100\% validation accuracy.

\medskip

\begin{figure}[t]
\centering
   \includegraphics[scale=0.45]{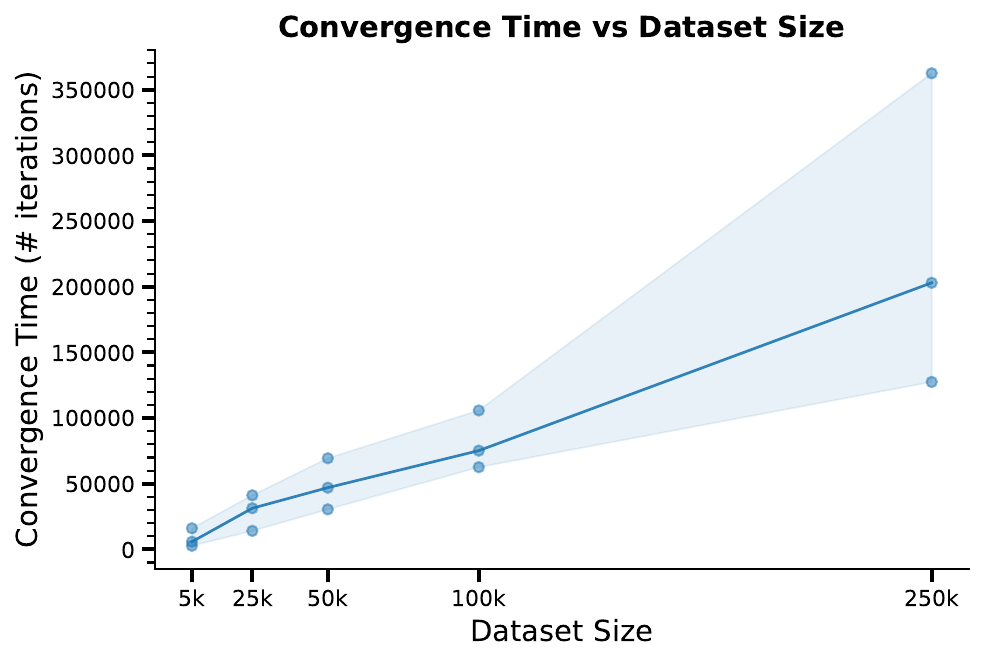}
	\caption{The number of computational steps till convergence vs size of training data for Transformers trained on $\sparn{40}{4}$. The curve depicts the median number (along with min/max) of iterations with a batch size of 500 across 50-100 successful runs for each dataset size. Refer to Section \ref{app:parities} for details. } \label{fig:convtime}
	
\end{figure}

\noindent \textbf{Convergence time vs Sample Size.} For Transformers trained on $\spar$, we conduct experiments to compare the number of computational steps required to successfully learn $\spar$ with the size of the dataset it is trained on. We consider length $n$=40 and $k$=4, and create datasets of five different sizes (\{5k, 25k, 50k, 100k and 250k\}). For each dataset, we train a Transformer of depth 2 and width 128 across 100 different initializations with learning rate $\in \{0.0001, 0.0005\}$ and with batch size 500. We consider each iteration as a computational step. We report the median, minimum and maximum steps for each dataset in Figure \ref{fig:convtime}. We find that neural networks such as Transformers can successfully learn $\spar$ with relatively small number of computational steps on small-sized training sets. It is perhaps surprising that for $\spar$ with $n$=40 and $k$=4, Transformers can successfully generalize well with less than $20000$ computational steps on over 75 out of 100 runs. 

\medskip

\medskip

\noindent \textbf{Phase Transitions.} As reported in \citet{hiddenprogressdl}, we observe phase transitions on Parity tasks where the training and validation accuracies do not change for a large number of training iterations and then abruptly reach near-perfect accuracy in a few iterations (see Figure \ref{fig:san_sparity_n}). This phenomenon was observed for feedforward networks (FFNs) and Transformers in \citet{hiddenprogressdl} and theoretically explained for ReLU FFNs trained with SGD. We observe another such behaviour for LSTMs on $\parn{n}$ (see Figure \ref{fig:rnn_parity} for training curves on $\parn{30}$). For both LSTMs and Transformers, we were unable to get them to generalize well with SGD on either $\spar$ or standard $\pari$. Both the architectures seem to succeed with the Adam optimizer \citep{adam}.

\medskip

\noindent \textbf{Grokking.} Another interesting phenomenon we observe in Transformers is that in some cases the training accuracies start increasing gradually with no change in the validation accuracy. After some iterations, the validation accuracy increases and matches the training accuracy. We reliably observed this phenomenon while training Transformers with absolute positional encodings across training sets of various sizes (see Figure \ref{fig:san_grok}) and while training with learnable encodings on small-sized training sets (see Figure \ref{fig:san5k_grok}). Similar observations for grokking \citep{grokking} were made in \citet{hiddenprogressdl} for ReLU FFNs trained on small-sized training sets. 

\subsection{Experiments on Variable Length Inputs}\label{app:varlen}

 We conducted some additional experiments on tasks with variable length inputs. These tasks are simple extensions of sparse parities and majorities to variable length input and have (in an informal sense) low sensitivity. 

\textbf{Task.} Let $\vparn{n}{k}$ denote the extension of $\sparn{n}{k}$ to variable length sequences. A function in $\vparn{n}{k}$ is defined over sequence of $\{0,1,2\}$ where the total number of $0$s and $1$s are exactly $n$, along with $k$ relevant indices which determine the label. The input distribution is such that there could be token $2$ between any zeros and ones with some probability. The tokens $2$ however do not influence the output of the function and are merely constructed to vary the input lengths. The label is determined by removing all the tokens $2$ from the input and applying the regular $\sparn{n}{k}$ over the remaining string over $\hamn$.

For illustration, for $f_S \in \sparn{4}{2}$, where $S=\{1,3\}$, for an input `$1001$', the function $f_S(\underline{1}0\underline{0}1)=1$ since the number of $1$s in position 1 and 3 is odd. For a similar function $\hat{f}_S \in \vparn{4}{2}$, here are some examples on various inputs: $\hat{f}_S(2\underline{1}022\underline{0}2221)= \hat{f}_S(\underline{1}20\underline{0}2212)= \hat{f}_S(\underline{1}2202\underline{0}21222)= f_S(\underline{1}0\underline{0}1)=1$. The function $\varmaj{n}{k}$ is defined similarly, where it takes an input string over $\{0,1,2\}$ and the label is determined by removing all $2$s and applying regular $\majnk{n}{k}$ on the remaining string over $\hamn$.

\medskip

\textbf{Results.} Contrary to the fixed length setting, we observe that both Transformers and LSTMs perform similarly on these tasks. For both $\vparn{n}{k}$ and $\varmaj{n}{k}$ we experiment with various mean lengths and variances with $k=5$. The general behaviour is that both Transformers and LSTMs generalize well when the tasks over short sequences ($<40$ for $\vparn{n}{k}$ and $<100$ for  $\varmaj{n}{k}$). However as the lengths of the input go beyond that, both architectures do not generalize well. 

In comparison to LSTMs, Transformers only performed better when the variance of the lengths of the inputs was very low. An interesting observation about Transformers is that they only seemed to generalize well with positional masking (also referred to as causal masking) along with positional encodings. Their performance was notably worse with only positional encodings (learnable or absolute). 

These results do not support the hypothesis posed in Section \ref{sec:intro} and we intend to explore this further in the future.

	

\section{Fitting Randomly Labelled Data}\label{app:rand_label}

We conduct some experiments to examine the ability of LSTMs and Transformers to fit random noise. The capacity of a class of functions to fit random noise is often theoretically measured as its Rademacher complexity. Given the incredible expressive power of neural networks, measures such as Rademacher complexity lead to vacuous generalization bounds. One assumption was that, despite their capacity, deep neural networks trained with gradient-based methods can only learn a small subset of such functions. The work of \citet{zhang2021understandingrethinking} demonstrated that large feedforward-like networks trained with gradient-based methods are able to fit random noise on image datasets. We conduct similar experiments to evaluate the ability of sequence models to fit noise on text data. We consider the SST dataset \citep{sst} as used in the GLUE benchmark. The training data contains approximately 65k samples and we label each sample either $+1$ or $-1$ randomly (with probability 1/2 each).

\begin{figure}[ht]%
	\centering
	\subfloat[\centering Transformers]{{\includegraphics[scale = 0.24, trim=0 0 5 5, clip]{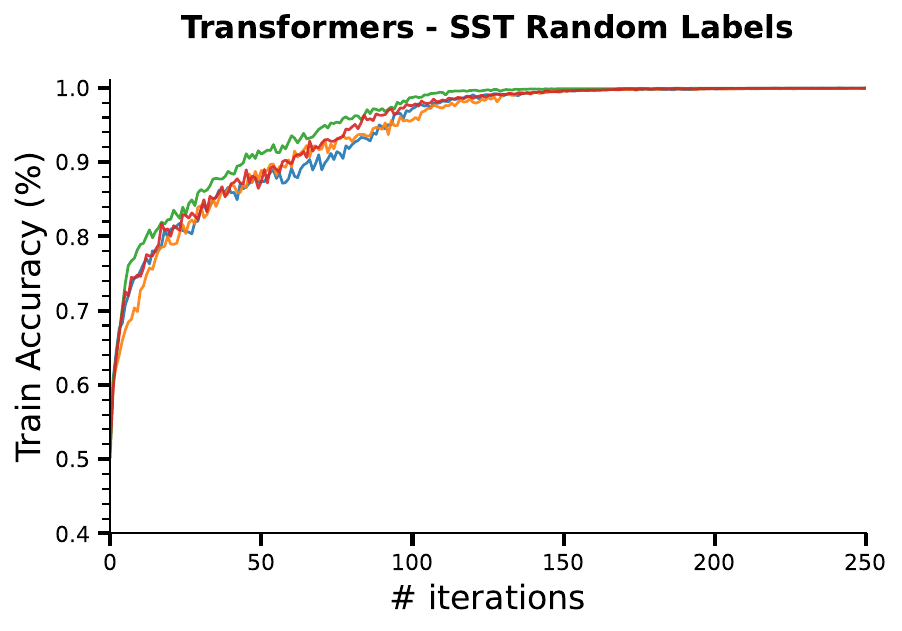} \label{fig:san_noise} }}%
	\subfloat[\centering LSTMs]{{\includegraphics[scale = 0.24, trim=0 0 5 5, clip]{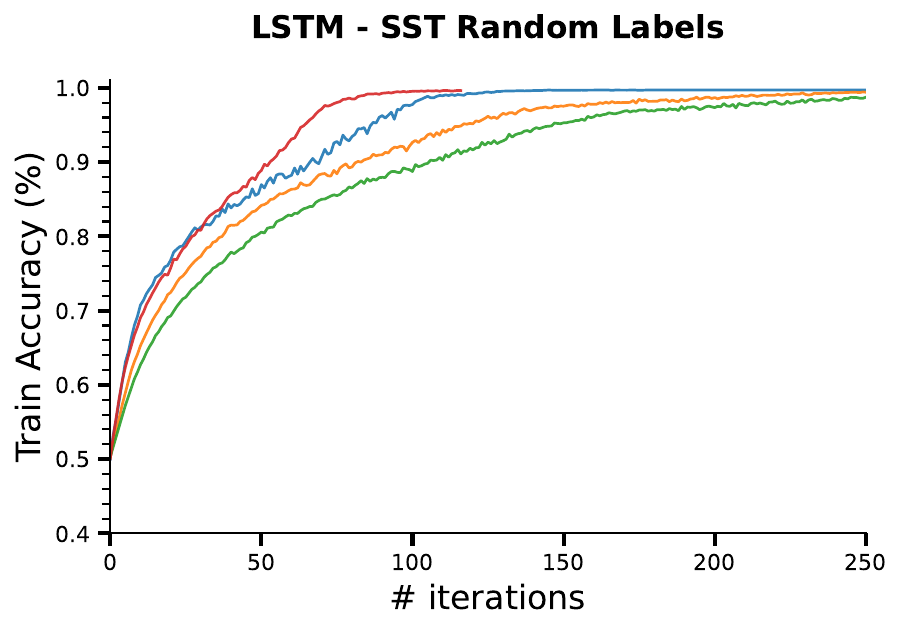} \label{fig:rnn_noise} }}
	
	\caption{Training curves of Transformers and LSTMs when trained on SST dataset with randomly assigned labels. Both architectures conveniently fit the entire dataset although they take much longer in comparison to when trained with original labels.}%
	\label{fig:sst_noise}%
\end{figure}

Figure \ref{fig:sst_noise} depicts the training curves for Transformers and LSTMs. We find that both the models are able to conveniently fit the training set near-perfectly. For both the models, the training takes significantly more number of iterations/epochs in comparison to training on the original dataset with true labels which only takes a few epochs.

\begin{table*}[t]
		\normalsize{\centering
	\begin{tabular}{P{12em}P{13em}}
		\toprule
		\textbf{Hyperparameter} & \textbf{Bounds} Transformer | LSTM\\
		\midrule
		D\_model/Hidden Size & [16, 128] | [8, 256]\\
		Heads & [4, 8 ] \\
		Number of Layers & [1, 6] | [1, 6]\\
		Learning Rate & [1e-2, 1e-5]\\
		Position Encoding Scheme & [Learnable, Absolute]\\
		\bottomrule
	\end{tabular}
	\caption{\label{tab:hyps} Different hyperparameters and the values considered for each of them. Note that certain parameters like Heads and Position Encoding Scheme are only relevant for Transformer-based models and not for LSTMs. }
}
\end{table*}

\section{Implementation Details}\label{app:implementation}

Our implementation of Transformer is based on \citet{rush-2018-annotated}. For various recurrent models such as RNNs, GRUs, and LSTMs, we use PyTorch's standard implementation \citep{pytorch}.

 For each dataset, we extensively tune across several hyperparameters and report the results based on the best-performing models. Table \ref{tab:hyps} lists the hyperparameters used for tuning the models for Boolean function experiments in Section \ref{sec:bool_exp}. We use a grid search procedure to tune the hyperparameters. The models were trained with cross-entropy loss. For all our results, we used Adam Optimizer and tuned the learning rates. We also tried SGD with weight decay but could not get either Transformers or LSTMs to perform well on parities, sparse parities, or random k-sparse functions. 

 \textbf{Compute.} All our experiments were conducted using 16 NVIDIA Tesla V100 GPUs each with 16GB memory. Since the datasets are synthetic and relatively smaller than practical datasets, most training runs took 10-30mins on a single GPU. The larger expenditure was on tuning LSTMs to find whether any of the hyperparameters succeed. Some experiments with LSTMs on $\sparn{40}{4}$ for over 100k steps took 3 hours and across multiple hyperparameters took $\approx$ 400 GPU hours in total. The experiments conducted for Figure \ref{fig:convtime} took similar amount of GPU hours ($\approx$ 300). The rest of the experiments took less than 10\% of this time. The experiments conducted in Section \ref{sec:sensi_exp} with random models were not as compute intensive.

\section{Additional Related Work}\label{app:relwork}

\textbf{Formal Languages and Recurrent Models.} For recurrent models, analysis on formal languages dates back to a few decades ago (see \citet{kolen2001field}). Several works have examined the ability of RNNs and LSTMs to recognize various context-free and counter languages \citep{gers2001lstm, weiss2018practical,suzgun2018evaluating}, the most prominent one being the Dyck\footnote{Checking whether a sequence of brackets is well-balanced.} languages \citep{skachkova-etal-2018-closing, bhattamishra-etal-2020-practical}.  Connections between RNNs and finite state automata have been explored for a long time \citep{RNNFSTgoudreau1994first,korsky2019computational}. Prior works have also sought to extract finite state automata from recurrent models trained on regular languages (see \citet{Comparative-Giles} for a survey). Connections between LSTMs and counter automata have also been established empirically \citep{suzgun2019lstm} and theoretically \citep{ merrill2020formal}. More recently, multiple works have investigated the ability of Transformers to recognize various regular, context-free \citep{ebrahimi2020can, yao-etal-2021-self,bhattamishra-etal-2020-practical}, and mildly context-sensitive languages \citep{wang2021evaluating}.

\textbf{Neural Networks and Parities.} Prior works can be divided into two categories. One set of works focuses on the Parity language containing strings of arbitrary length that can be represented by a 2-state DFA. The other set examines the $\sparn{n}{k}$ problem with strings over $\hamn$ where the output depends on a subset of bits. The $\sparn{n}{k}$ problem has been widely studied in learning theory and has several well-understood properties. On the $\sparn{n}{k}$, \citet{hiddenprogressdl} theoretically analyze the ability of feedforward-like networks trained with SGD and conduct experiments with various architectures including Transformers. Some of our results corroborate their findings and we empirically explore the phenomenon further. Preliminary experiments along this direction were also explored in \citet{edelman2022inductiveksparse} where they showed that Transformers can efficiently express $\kspar$ functions.

Since variable length Parity can be represented by a 2-state DFA, small-sized RNNs can efficiently represent them and several works have found LSTMs to generalize well when tested empirically \citep{schwarzschild2021can}. On the other hand, Transformers are limited in their ability to express such a language \citep{hahn-2020-theoretical}. While they can express them for bounded lengths \citep{chiang-cholak-2022-overcoming}, they have been found to struggle when tested empirically \citep{bhattamishra-etal-2020-ability,deepminddeletang2022neural,chiang-cholak-2022-overcoming}. \citet{anil2022exploring} explore length generalization abilities of large language models on Parity. Finding $\pari$ and other languages by uniformly initializing weights was also explored in the 1990s (see Chap. 13 \citet{kolen2001field}) for older versions of recurrent architectures which are not used anymore.

\end{document}